\documentclass[runningheads]{llncs}

\usepackage[utf8]{inputenc}

\usepackage{eccv}

\usepackage{hyperref}
\usepackage{orcidlink}

\usepackage{eccvabbrv}
\usepackage[accsupp]{axessibility}  

\usepackage{booktabs}
\usepackage{hyperref}
\usepackage{multirow}
\usepackage{adjustbox}
\usepackage{xspace}
\usepackage{graphicx}
\usepackage[dvipsnames]{xcolor}

\usepackage{comment}
\usepackage{csquotes}
\usepackage{wrapfig}
\usepackage{lscape}
\usepackage{pifont}
\usepackage{environ}
\usepackage{array}  
\usepackage[normalem]{ulem}
\usepackage{adjustbox}
\usepackage{multirow}
\usepackage{subcaption}
\usepackage{bm}

\usepackage{svg}
\usepackage{tikz}
\usetikzlibrary{matrix, positioning, calc}
\usepackage{array}

\usepackage{listings}
\usepackage{algorithm}
\usepackage{float}

\usepackage[varqu]{zi4}
\usepackage{tcolorbox}
\usepackage[frozencache,cachedir=.]{minted}
\usepackage{amsmath}
\usepackage{amssymb}

\definecolor{hlcol}{RGB}{255,200,200}

\newcommand{\mintedpseudocode}[1]{
    \begin{tcolorbox}[arc=1pt,colback=black!3!white,colframe=black!16!white,boxrule=0.3mm,before skip=1mm,after skip=1mm,top=.5mm,bottom=.5mm,left=1mm,right=1mm]
         \inputminted[escapeinside=||,mathescape=true,fontsize=\footnotesize,linenos]{python}{#1}
    \end{tcolorbox}
}

\newfloat{mintedalgorithm}{thp}{lop}
\floatname{mintedalgorithm}{Algorithm}

\definecolor{sig_gray}{rgb}{0.96,0.96,0.96}    
\definecolor{sig_red}{rgb}{0.6,0.1,0.1}        
\definecolor{sig_num}{rgb}{0.2,0.6,0.6}        

\lstdefinestyle{siggraph_style}{
    language=Python,
    backgroundcolor=\color{sig_gray},
    basicstyle=\ttfamily\small,
    keywordstyle=\bfseries\color{sig_red},
    commentstyle=\color{gray},
    frame=trbl,
    frameround=tttt,
    rulecolor=\color{gray!50},
    xleftmargin=3pt,    
    xrightmargin=3pt,
    aboveskip=10pt,     
    belowskip=10pt,     
    mathescape=true,
    escapechar=!,
    showstringspaces=false,
    gobble=4            
}

\newcommand{\loss}{g}
\newcommand{\image}{I}
\newcommand{\bsdf}{f_s}
\newcommand{\unitsphere}{S^2}
\newcommand{\param}{\pi}
\newcommand{\paramv}{\bm{\pi}}

\newcommand*\diff{\mathop{}\!\mathrm{d}}

\DeclareMathOperator*{\argmin}{arg\,min}

\newcommand{\bomega}{\bm{\omega}}

\newcommand{\vL}{\mathbf{L}}
\newcommand{\vx}{\mathbf{x}}

\newcommand{\sv}{\mathbf{s}}

\usepackage{xcolor}

\usepackage{soul}
\definecolor{junglegreen}{rgb}{0.113, 0.639, 0.5}

\begin{document}

\title{Differentiable Polarized Path Tracing}



\author{
Pramod Rao\inst{1,2,3} \and
Jérémy Riviere\inst{4} \and
Xilong Zhou\inst{1,2} \and
Abhijeet Ghosh\inst{4} \and
Abhimitra Meka\inst{4} \and
Thabo Beeler\inst{4} \and
Marc Habermann\inst{1,2,3}\and \\
Christian Theobalt\inst{1,2,3} \and
Delio Vicini\inst{4}
}

\authorrunning{P.~Rao et al.}

\institute{
Max Planck Institute for Informatics \and
Saarland Informatics Campus \and
VIA Research Center \and
Google
}

\maketitle


\vspace{-1cm}
\begin{figure}
    \centering
    \includegraphics[width=\textwidth]{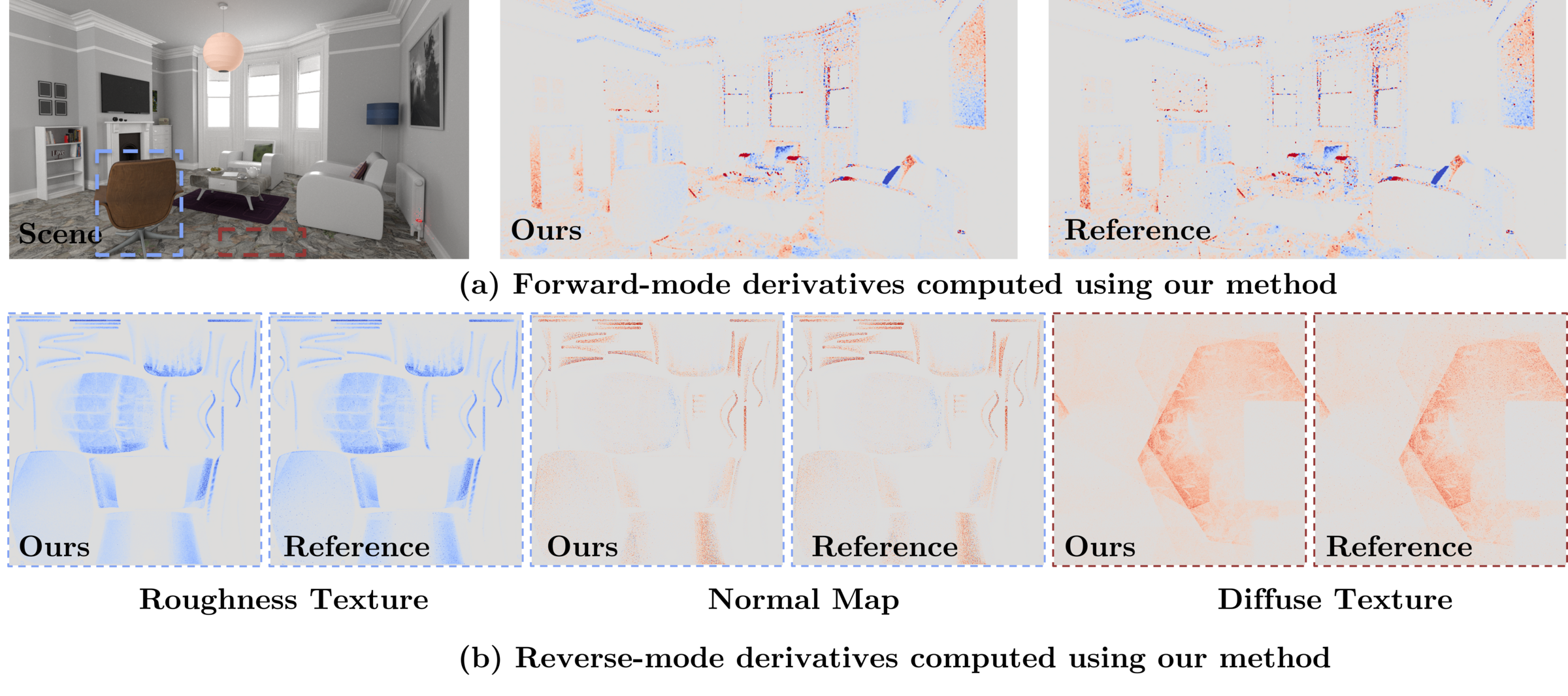}
    \vspace{-0.4cm}
    \caption{Polarization plays an important role in differentiable inverse rendering. \textbf{(a)} Parameters that control polarization effects such as a linear polarizer rotation $\theta$ can produce significant gradient contributions, but existing differentiable path tracing methods struggle to compute these gradients efficiently and accurately. We introduce a caching mechanism for polarized path replay that avoids inverting rank-deficient Mueller operators. \textbf{(b)} Our method produces reverse-mode gradients with respect to scene parameters that match conventional automatic differentiation (AD), while using substantially less memory. Our algorithm improves runtime relative to existing unbiased baselines and yields substantially lower gradient error.}
    \label{fig:teaser}
\end{figure}
\vspace{-1cm}
\begin{abstract}

Physically based differentiable rendering has proven to be a powerful tool for inverse rendering problems (e.g., 3D reconstruction, reflectance estimation, lighting estimation). However, most existing methods operate solely on radiometric intensity, discarding valuable polarization cues that constrain scene geometry and material properties. While forward simulation of polarized light is well-defined via Mueller-Stokes calculus, extending reverse-mode differentiation to this domain presents significant challenges. The rank-deficient nature of common polarimetric operators, such as linear polarizers and diffuse reflections, violates the invertibility assumptions of standard gradient estimators like path replay backpropagation and results in numerical instability. We address this by proposing a robust, polarization-aware differentiable path tracing method. Our approach estimates unbiased gradients through a combination of path replay and local caching. This formulation enables efficient and stable optimization of material and lighting parameters in complex scenes, broadening the applicability of physically based inverse rendering.
Project page: \url{https://vcai.mpi-inf.mpg.de/projects/DPPT/}


\end{abstract}

\section{Introduction}
Polarimetric imaging offers valuable cues for inverse problems such as 3D reconstruction~\cite{kadambi2015polarized, ba2020deep, lei2022shape, han2025polgs, atkinson2005multi}, appearance estimation~\cite{ma_polarization, Riviere2017, Baek2018, Deschaintre2021}, and intrinsic decomposition~\cite{wolff1989using, wolff1990polarization}. Capturing information about the polarization state of light helps constrain reconstructions to disambiguate material or surface properties that may appear similar in intensity-only images~\cite{wolff1989using}.

Recently, \emph{inverse rendering} approaches have gained popularity to tackle 3D reconstruction problems~\cite{mildenhall2020nerf, kerbl20233d, NimierDavidVicini2019Mitsuba2, Weier2025PracticalInverse}. Specifically, \emph{physically based differentiable rendering (PBDR)} methods optimize scene parameters while accounting for the intricate interaction of light with scene geometry and materials~\cite{gkioulekas2016evaluation, Li18EdgeSampling, Zhang2019DTRT, NimierDavidVicini2019Mitsuba2}. Using reverse-mode differentiation, these methods can jointly optimize very high-dimensional parameter vectors (e.g., textures or mesh vertices). The practical application of these methods can be challenging due to local minima and ambiguities. In this work, we combine polarimetric measurements with PBDR. By using these richer observations, we enable better identification of material parameters in inverse rendering and improve robustness.

The combination of PBDR with polarization information is non-trivial: While PBDR methods offer a powerful and general framework, simply applying reverse-mode automatic differentiation~\cite{Griewank2008} to path tracing~\cite{kajiya1986} quickly exhausts the available system memory~\cite{NimierDavid2020Radiative, vicini2021PathReplay}. \emph{Path replay backpropagation~(PRB)}~\cite{vicini2021PathReplay} addresses this by leveraging local numerical invertibility to efficiently compute gradients using a constant amount of memory and linear complexity. However, simulating polarized light transport typically uses \emph{Mueller-Stokes calculus}~\cite{collet2005}. This means scalar radiance values are replaced by 4D Stokes vectors, and interactions of light with surfaces require multiplication with $4 \times 4$ Mueller matrices. Even for simple scenes, the Mueller matrices are often singular and thus violate PRB's local invertibility assumptions. Therefore, a direct application of PRB leads to unstable or severely biased gradient estimators.

We first identify these numerically problematic cases and the scenarios where standard PRB fails. We then propose a general, efficient, and easy-to-implement algorithm that produces unbiased gradients (\cref{fig:teaser}). The key idea is to extend PRB by selectively storing a small amount of additional information in local memory. This is significantly less memory-intensive than conventional automatic differentiation while resolving numerical issues, which enables fully differentiable polarized path tracing. When applying this method to material reconstruction tasks, the additional cues from polarimetric imaging effectively reduce problematic local minima that hinder  unpolarized inverse rendering. In this paper, we focus on the continuous \emph{interior} integral of the differentiable light transport, including the handling of geometric discontinuities and shape reconstruction under polarization. In summary, our contributions are:




\begin{itemize}
    \item An analysis of why standard PRB does not directly generalize to polarized light transport, identifying rank-deficient Mueller operators as the source of ill-defined or unstable replay cancellation.

    \item To our knowledge, the first practical framework for differentiable polarized path tracing along long light paths, combining cached suffix replay with a hybrid variant to avoid unreliable Mueller-matrix inversion.

    \item An experimental validation demonstrating that polarized PBDR improves material, texture, normal recovery, and geometry reconstruction compared to unpolarized baselines while producing gradients consistent with Conv AD.
\end{itemize}

\section{Related Work}
\subsection{Polarimetric Reconstruction} 

Polarization has a long history in computer vision and graphics, dating back to the 1970s when polarization was established as a useful signal for shape understanding \cite{koshikawa1979polarimetric,wolff1990surface,boult1991physically}, material classification \cite{wolff1989material,wolff1990polarization}, and the separation of diffuse and specular reflection \cite{wolff1989using}. 
The seminal work of Boult \& Wolff \cite{boult1991physically} establishes the theoretical framework for Shape from Polarization (SfP) for specular reflection via the Transmitted Radiance Sinusoid (TRS), measured by rotating a linear polarizer in front of a camera at 3–4 angles\footnote{Typically $0^\circ$, $\pm45^\circ$, and $90^\circ$. A polarization camera may also be used, where the Bayer pattern is replaced by micro-polarizers at different orientations per pixel.}. The phase of the TRS is related to the surface normal up to an ambiguity of $180^\circ$ in azimuth. Subsequent works address this ambiguity using multi-view constraints \cite{atkinson2005multi,miyazaki2012polarization}, fusion with coarse geometry \cite{kadambi2015polarized}, or deep learning approaches \cite{ba2020deep,lei2022shape}. While most SfP methods focus on dielectric specular materials, extensions to transparent objects \cite{miyazaki2003polarization} and diffuse surfaces \cite{atkinson2006recovery} have also been proposed. 
Another interesting application of polarization for scene understanding is diffuse-specular separation, where specular surfaces preserve the polarization properties of incident polarized illumination while diffuse surfaces depolarize light. Ma~\etal~\cite{ma_polarization} introduce cross and parallel polarized spherical gradient illumination to recover diffuse and specular albedo as well as surface normals. Ghosh~\etal~\cite{ghosh_multiview_polarization} further extend this setup to multi-view polarization capture.
Researchers also investigate methods to reduce capture complexity for polarimetric reflectance acquisition. Baek \etal~\cite{Baek2018} propose a near co-axial projector-camera ellipsometry setup for acquiring shape and spatially varying surface reflectance, combining inverse rendering with a polarimetric BRDF (p-BRDF) modeling both specular and diffuse polarization effects. Baek \etal~\cite{Baek2020} further extend this to spectral measurements and provide a database of isotropic polarimetric BRDFs. Hwang \etal~\cite{Hwang2022} simplify capture further by using sparse ellipsometry with a hand-held polarized flash camera for shape and SVBRDF reconstruction. Deschaintre \etal~\cite{Deschaintre2021} leverage deep learning frameworks for 3D shape and material reconstruction. Riviere \etal~\cite{Riviere2017} propose a method to acquire material reflectance properties through multi-view polarized images, which was later extended by \cite{nogue2022} to single-view material acquisition. Polarization techniques are also widely used for complex facial appearance capture \cite{polarface2024, Lin2025}.

Polarization has also been integrated with neural fields to improve geometry reconstruction and challenging material acquisition. Han \etal~\cite{han2024nersp} address the specific problem of reconstructing reflective objects from sparse views. By incorporating a p-BRDF~\cite{Baek2018} and enforcing normal consistency derived from the Angle of Polarization (AoP) cues, NeRSP significantly reduces shape-radiance ambiguity for sparse captures. Moving towards higher physical fidelity, Wanaset \etal~\cite{wanaset2025neural} propose a method that fits a polarized neural field directly to raw, mosaiced sensor data. More recently, Han \etal~\cite{han2025polgs} explicitly incorporate polarimetric constraints into 3D Gaussian Splatting~\cite{kerbl20233d}. They utilize a Gaussian Surfel representation to better approximate surface geometry and employ a p-BRDF~\cite{Baek2018} module to separate specular and diffuse components, allowing high-fidelity reconstruction of shiny objects from sparse inputs. Concurrently, Guo \etal~\cite{guo2025polargs} focus on ambiguity-free reconstruction. They leverage the Degree of Linear Polarization (DoLP) to detect photometrically ambiguous regions (e.g., textureless or perfectly specular areas) and apply a polarization-guided geometric correction to align Gaussian normals with physical constraints.

While these methods demonstrate promising reconstruction results, they primarily use polarization cues as additional supervision signals during optimization, without addressing the underlying challenge of differentiating through the polarized light transport governed by Mueller-Stokes calculus. In contrast, our work addresses the fundamental rank-deficiency of polarimetric operators and extends PRB to support polarization-aware differentiable path tracing, enabling efficient and stable inverse rendering.
\vspace{-1.2em}
\subsection{Physically Based Inverse Rendering}
Physically based inverse rendering methods differentiate (unbiased) Monte Carlo path tracing~\cite{gkioulekas2016evaluation, Li18EdgeSampling,NimierDavidVicini2019Mitsuba2,Zhang2019DTRT,Weier2025PracticalInverse}. Many works in this space focus on differentiating visibility discontinuities, as required for geometry optimization~\cite{Li18EdgeSampling, Zhang2019DTRT, bangaru2020warpedsampling, Zhang2023Projective, Xu2024mcmc}. On the other hand, Stam~\cite{DBLP:journals/corr/abs-2006-15059}, Nimier-David \etal~\cite{NimierDavid2020Radiative} and Vicini \etal~\cite{vicini2021PathReplay} present approaches to reduce the memory use for differentiating long light paths. Specifically, we extend path replay backpropagation~\cite{vicini2021PathReplay} to support polarization. Handling visibility discontinuities is orthogonal, and we leave this for future work. To the best of our knowledge, Yang \etal~\cite{Yang2026} is the only prior work demonstrating differentiable polarization rendering across multiple interactions. The authors demonstrate the use of a differentiable polarized light transport simulation for waveguide design. The implementation is specialized to waveguide design, and uses up to 70 GB of memory across four GPUs. Our method is aimed at general scenes, and carefully avoids storing large gradient checkpoints.
\section{Background}

\subsection{Inverse Rendering}

\paragraph{Problem Definition.} Our goal is to solve inverse rendering problems of the form: 
\begin{align}
    \paramv^* = \argmin_{\paramv} \loss(\image(\paramv)),
\end{align}
where $I$ is a rendered image as a function of a vector of scene parameters $\paramv$, and $\loss$ is a differentiable objective function.
To solve this problem using gradient descent, we need to efficiently compute derivatives of the rendering function $\image(\paramv)$. Without loss of generality, all the following derivations will be written in terms of a single pixel and a single parameter $\param$.

\paragraph{Rendering Equation.} Physically based rendering algorithms estimate the pixel color by applying Monte Carlo integration to the \emph{rendering equation}~\cite{kajiya1986}:
\begin{align}
    L_o(\vx, \bomega_o) = L_e(\vx, \bomega_o) + \int_{\unitsphere} \bsdf \left(\vx, \bomega_o, \bomega_i\right) L_i(\vx, \bomega_i) \diff \bomega_i^{\perp},
\end{align}
where $\vx$ is a position, $\bomega_o$ the outgoing direction, and $\bomega_i$ the incident direction. $L_o$, $L_i$, and $L_e$ denote outgoing, incident, and emitted radiance, respectively. The bidirectional scattering distribution function~(BSDF)~$\bsdf$ differentially relates incident to outgoing radiance. In unpolarized rendering, radiance is either modeled spectrally, or simply as RGB values. The standard representation to simulate polarization uses Stokes vectors and Mueller matrices (\cref{sec:polarization}).

\subsection{Path Replay Backpropagation}
Before expanding on polarized light transport, we review path replay backpropagation~(PRB)~\cite{vicini2021PathReplay}. The path replay algorithm allows computing (unbiased) derivatives of arbitrarily long light paths. For this, it exploits the local invertibility of the path tracer's light path contribution calculation. Our method is an extension of this algorithm to polarized light transport.


\vspace{1em}
\noindent
\begin{minipage}{\linewidth}
    \makeatletter\def\@captype{algorithm}\makeatother
    \mintedpseudocode{code/prb.py}
    \caption{\label{code:prb} Pseudocode for path replay backpropagation.}
\end{minipage}
\vspace{1em}

PRB computes gradients in three phases: 
\begin{enumerate}
    \item An initial \emph{primal} phase estimates the image pixels and is used to compute the derivative of the loss function with respect to the pixel intensities. 
    \item A decorrelated rendering pass traces new light paths and retains the estimated radiance per path.
    \item Finally, the \emph{adjoint} pass reuses the random seed from step 2, \emph{replaying} the same light paths. At each interaction, it differentiably evaluates the local contribution of the BSDF and emitters, and accumulates the gradients due to that particular interaction. 
\end{enumerate}
Notably, automatic differentiation is only used within the loop body and not across the iteration boundaries. This avoids memory-intensive checkpoints that retain the full loop state, as would be needed to differentiate a loop using automatic differentiation~\cite{Griewank2008}.

Algorithm~\ref{code:prb} provides pseudocode for the adjoint phase (step 3) of the algorithm. The pseudocode computes derivatives of the BSDF parameters. The inputs of the algorithm are the current ray, the resulting radiance estimate from step 2, and the gradient of the loss function with respect to the path radiance $\delta L$. This adjoint pass evaluates a vector-Jacobian product between the Jacobian $\partial L / \partial \paramv$ and the vector $\delta L$. The term $\beta$ denotes the \emph{path throughput} and $L_e$ is the emitted radiance. The \texttt{sample\_bsdf} function samples an outgoing direction proportional to the BSDF. Line 6 evaluates the local derivative of the current BSDF parameters. The function \texttt{backward(x)} ``backpropagates'' gradients of the computation \texttt{x}. PRB computes gradients selectively with respect to only some of the quantities within the loop body. For clarity, we therefore indicate the differentiated quantity in blue color, with any non-differentiated weights in black. We refer to \cite{vicini2021PathReplay} for an in-depth introduction to PRB.

\subsection{Polarized Light Transport}
\label{sec:polarization}
We now define notation and terminology related to polarized rendering.
For a comprehensive theory, please refer to Collet \etal~\cite{collet2005}, and Zeltner \etal~\cite{mitsuba2025polarization}.

\paragraph{Mueller-Stokes Calculus.} The standard approach to rendering with polarization is to use \emph{Mueller-Stokes calculus}. In this framework, the scalar radiance is replaced by 4D \emph{Stokes vectors}. A Stokes vector $\sv \in \mathbb{R}^4$ has components $s_0$ for intensity, $s_1$ distinguishes horizontal and vertical linear polarization, $s_2$ accounts for diagonal polarization, and $s_3$ for circular polarization. For unpolarized light, components $s_1$, $s_2$ and $s_3$ are zero. A \emph{Mueller matrix} is a $4 \times 4$ matrix that characterizes a linear operation on Stokes vectors. In polarized light transport, a BSDF evaluation returns a Mueller matrix. Various optical filters (e.g., a linear polarizer) can be modeled using their Mueller matrix.

\paragraph{Coordinate Conventions.} Polarized quantities are always defined with respect to a 3D coordinate system. Any operation (e.g., addition of two Stokes vectors) requires both quantities to be expressed with respect to the same coordinate frame. We implement our algorithms on top of Mitsuba~3~\cite{jakob2022mitsuba3}, which implements all the necessary coordinate conversions.

\paragraph{Path Contribution.} Implementing the Mueller-Stokes calculus requires modifying Algorithm~\ref{code:prb}. For simplicity, a practical implementation can represent all quantities using Mueller matrices~\cite{NimierDavidVicini2019Mitsuba2}. For that, Stokes vectors are represented as a Mueller matrix, where the first column contains the Stokes vector, and the other entries are zeros. Therefore, in a polarized path tracer, both the accumulated path contribution and throughput become $4 \times 4$ matrices. A key assumption of PRB is that \texttt{bsdf\_val} is non-zero, and hence invertible. In polarized rendering, \texttt{bsdf\_val} is a $4 \times 4$ Mueller matrix. For many polarization operators (e.g., optical filters), the Mueller matrix is singular (or close to it), as it maps the 4D Stokes space to the lower-dimensional subspace of the transmitted/reflected polarization state. Mathematically, such a lossy operation does not have a well-defined inverse. Therefore, the scalar inversion used by PRB does not directly carry over to the polarized setting.

\section{Method}

\subsection{Problem Statement}
We discuss the generalization of PRB to support polarized rendering. Our goal is to enable the computation of gradients with an arbitrary path depth while avoiding approximations and bias.

\paragraph{Pseudocode.} Algorithm~\ref{code:prb_polarized_basic} shows a basic implementation of PRB applied to polarization. The key difference from Algorithm~\ref{code:prb} is that all quantities are now Mueller matrices, and $M^{-1}$ denotes matrix inversion.

\paragraph{Limitations of Algorithm~\ref{code:prb_polarized_basic}.} In practice, the matrix $M$ is most often rank-deficient and non-invertible. For example, an ideal diffuse BSDF's Mueller matrix $M$ has only a single non-zero component $M_{00}$ (the albedo). All other entries are zero, and the matrix is therefore not invertible. Intuitively, a diffuse interaction loses information about the incident light's polarization state. Similarly, a linear polarizer filters out circular polarization, thus losing information about the last component of the Stokes parameters. On the other hand, some optical elements do produce invertible matrices. For example, a quarter waveplate simply rotates the incident polarization state by 90$^\circ$ around its axis, and does not lose any information.

\vspace{0.3em}
\noindent
\begin{minipage}{\linewidth}
    \makeatletter\def\@captype{algorithm}\makeatother
    \mintedpseudocode{code/prb_polarized_basic.py}
    \caption{\label{code:prb_polarized_basic} Pseudocode for polarized path replay backpropagation.}
\end{minipage}
\vspace{0.3em}

\paragraph{Analysis.} The examples above reveal that there are three different cases our algorithm ultimately needs to handle:

\begin{enumerate}
    \item \emph{Invertible elements}: Any BSDF with an invertible Mueller matrix can be handled by directly inverting $M$.
    \item \emph{Ideal depolarizers}: BSDFs that fully depolarize incident illumination (e.g., diffuse BSDFs) can be handled by standard PRB. Their derivatives do not require any information about the incident polarization state of light.
    \item \emph{Non-invertible polarization-aware elements}: Elements that consider polarization, but produce a non-invertible $M$ are not supported by the baseline algorithm.
\end{enumerate}%
The first two cases can already be handled within the framework of the basic Algorithm~\ref{code:prb_polarized_basic}, but the last case requires additional considerations. 

\subsection{Regularization Using Noise}
Vicini \etal \cite{vicini2021PathReplay} propose to handle a similar non-invertibility for \emph{attached} sampling by adding i.i.d. noise to the matrix diagonal. The key idea here is to add $u \cdot I_4$ to the matrix $M$, where $u \sim \mathcal{N}(0, \sigma)$ is randomly sampled. By using the same noise for both the initial path tracing as well as the adjoint phase, this technically does not introduce bias into the gradients~\cite{vicini2021PathReplay}. This regularization can help in some cases, but is not without issues: First, there is no guarantee that the resulting matrix is well-conditioned. Second, adding too much noise will deteriorate the quality of the derivatives. This method alone cannot produce sufficiently high-quality gradients, regardless of the choice of the noise magnitude $\sigma$, as shown in \cref{subsec:results_gradient_analysis}.

\begin{figure}[t]
    \centering
    \includegraphics[width=1\linewidth]{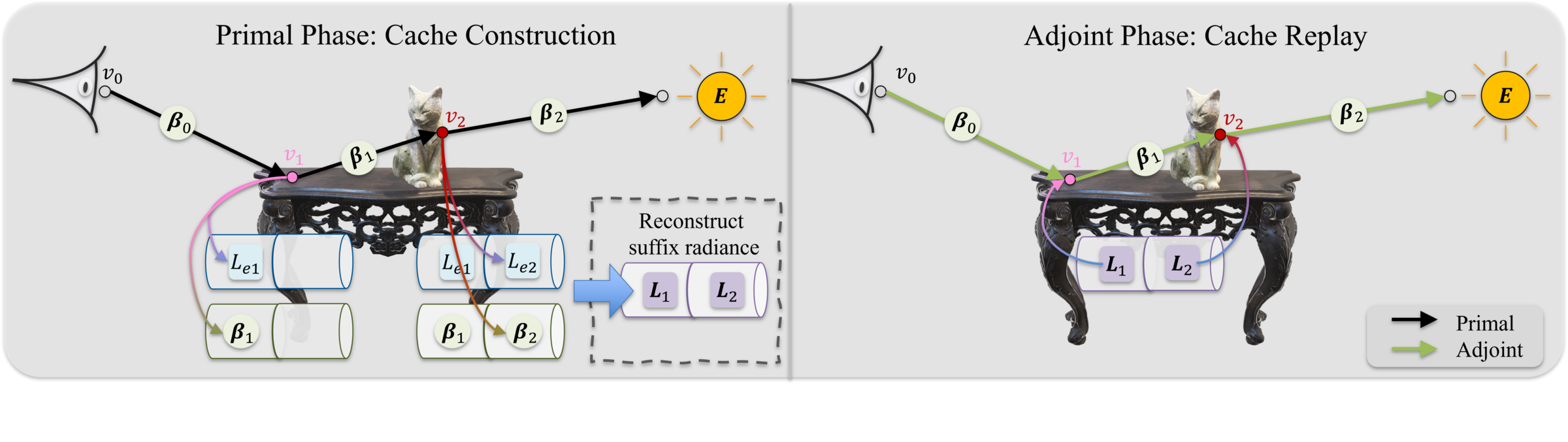}
    \vspace{-0.5cm}
    \caption{Example for Cached Suffix Replay on a light path of depth 2. The primal pass accumulates the full path contribution, tracks the detached BSDF contributions $\beta_i$, and constructs the cache of suffix contributions via a backward fold. The adjoint pass then directly re-loads the incident polarization state and multiplies it by the differentiated BSDF weight. This enables unbiased gradient estimation without any matrix inversion.}
    \label{fig:method_illustration}
\end{figure}
\subsection{Cached Suffix Replay}
\label{subsec:method_cached_suffix}

The basic polarized extension of PRB (Algorithm~\ref{code:prb_polarized_basic}) follows the same replay logic as in the scalar setting. During the adjoint pass, it ``removes'' the detached BSDF contribution by multiplying with the inverse of the sampled transport factor. In polarized light transport, this factor is a Mueller matrix $M$, which is frequently non-invertible, and thus results in incorrect derivatives. Our alternative avoids inversion by caching a compact summary of the \emph{suffix} of each sampled path. We illustrate the core idea using an example light path in \cref{fig:method_illustration}.

\vspace{0.1cm} 
\noindent
\begin{minipage}{\linewidth}
    \makeatletter\def\@captype{algorithm}\makeatother
    \mintedpseudocode{code/prb_polarized_cached.py}
    \caption{\label{code:prb_polarized} Pseudocode: polarized path replay backpropagation with caching.}
\end{minipage}
\vspace{-0.8cm} 

\paragraph{Algorithm.} We implement this cached path replay with two passes (Algorithm~\ref{code:prb_polarized}). The primal pass stores per-vertex emission and throughput factors in local arrays and reconstructs suffix radiance by a backward fold. During the adjoint replay, instead of canceling the detached contribution via $M^{-1}$, we directly look up the cached suffix radiance $L_i=\vL[i+1]$. We then compute the local gradient term using the replayed throughput and a differentiable BSDF evaluation. This removes the need for inverting rank-deficient Mueller operators while producing unbiased gradients. Unlike conventional automatic differentiation, we do not store the complete loop state, but instead retain path replay to recompute surface interaction structures, BSDF values and next-event estimation. This offers a more favorable memory-compute tradeoff, and scales to scenes of high complexity, as demonstrated in the \cref{sec:results}.


\subsection{Hybrid Cached Replay}
\label{subsec:method_hybrid}

Cached Suffix Replay stores suffix radiance at every path interaction. This is efficient for the path depths used in most of our inverse-rendering experiments, but the local cache size still grows with the maximum path depth. To reduce this storage for deeper paths, we introduce a hybrid variant that combines block-wise caching with recursive recomputation.

The hybrid method stores cached suffix values only at checkpoint intervals. Let \(k\) denote the checkpoint interval. When \(k=1\), the method reduces to the fully cached variant described above. Larger values of \(k\) store fewer cached states and therefore reduce memory usage, but require additional work during replay for vertices that lie between two checkpoints. At such vertices, the adjoint pass cannot directly load the suffix radiance and must recover it either by local inversion or by recursively tracing until the next checkpoint.

The decision is made using the detached block throughput \(\bar{\beta}\), which summarizes the polarized transport accumulated since the last checkpoint. We first evaluate the depolarization index \(DI(\bar{\beta})\). If \(DI(\bar{\beta}) < \gamma\), the accumulated transport is treated as effectively depolarizing and we use the scalar PRB update. Otherwise, we retain the polarized update and test whether local inversion is numerically admissible. The high-level replay rule is:
\begin{align}
\label{eq:hybrid_decision}
\text{suffix recovery} =
\begin{cases}
\text{scalar PRB update}, &
DI(\bar{\beta}) < \gamma, \\
\bar{\beta}^{-1}L, &
DI(\bar{\beta}) \ge \gamma \ \text{and}\ |\det(\bar{\beta})| > \varepsilon, \\
\text{recursive recomputation}, &
\text{otherwise},
\end{cases}
\end{align}
where \(L\) is the cached suffix at the next checkpoint. The threshold \(\gamma\) controls when polarized replay is retained, while \(\varepsilon\) avoids unstable inversions of near-singular Mueller operators. This yields a memory-reduced variant of our method for larger path depths. Full pseudocode and extended benchmarks are provided in the supplemental material.
\section{Results}
\label{sec:results}

In this section, we report correctness tests, practical applications of polarization, and performance evaluations of our method.
We ran all evaluations on a workstation equipped with an NVIDIA GeForce RTX 4090 (24~GiB). To our knowledge, aside from Mitsuba's conventional automatic differentiation, our method is the first to support full differentiable polarized path tracing. We therefore compare against polarized extensions of PRB \cite{vicini2021PathReplay} and RB \cite{NimierDavid2020Radiative}, which we denote P-PRB and P-RB, respectively.
While these extensions are not part of the original works, we discussed the implementation details with the authors and validated their correctness.

\subsection{Gradient Analysis}
\label{subsec:results_gradient_analysis}
\begin{figure*}[t]
\centering
\newcommand{\colwidth}{0.185} 

\begin{tikzpicture}[
    every node/.style={inner sep=0,outer sep=0},
    col_label/.style={font=\sffamily\bfseries\small, anchor=south, yshift=1mm},
    row_label_left/.style={
      font=\sffamily\bfseries\tiny, 
      rotate=90,
      anchor=south,
      yshift=1mm,
      text width=18mm,     
      align=center
    },
    row_label_right/.style={
      font=\sffamily\bfseries\tiny, 
      rotate=270,
      anchor=south,
      yshift=1mm,
      text width=18mm,     
      align=center
    },
    rebox/.style={
        font=\sffamily\bfseries\tiny,
        text=white,
        anchor=south east,
        xshift=-0.8mm,
        yshift=0.6mm,
        fill=black,
        fill opacity=0.45,
        text opacity=1,
        rounded corners=0.6pt,
        inner sep=1.2pt
    },
]

\matrix (grid) [
    matrix of nodes, 
    nodes={anchor=center}, 
    column sep=1mm, 
    row sep=1mm
] {
    \includegraphics[trim={1cm 1cm 4cm 1cm}, clip, width=\colwidth\linewidth]{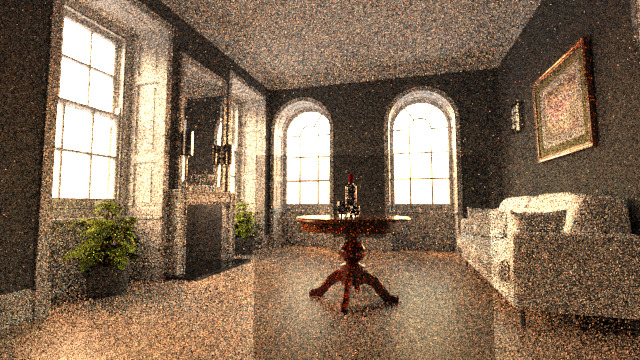} &
    \includegraphics[trim={1cm 0cm 4cm 0cm}, clip, width=\colwidth\linewidth]{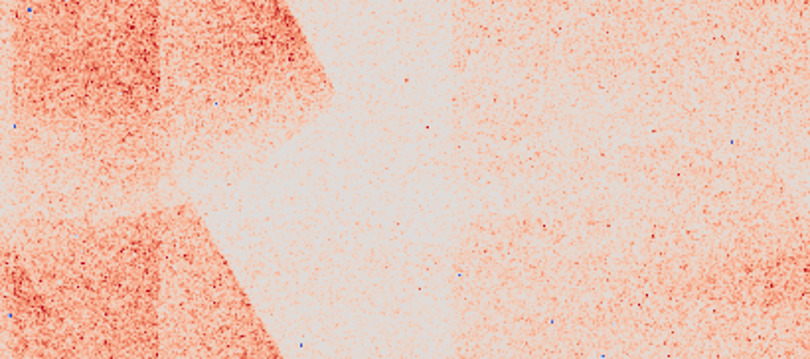} &
    \includegraphics[trim={1cm 0cm 4cm 0cm}, clip, width=\colwidth\linewidth]{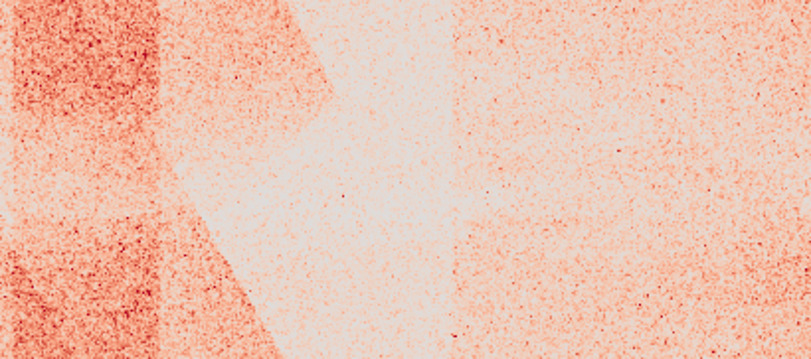} &
    \includegraphics[trim={1cm 0cm 4cm 0cm}, clip, width=\colwidth\linewidth]{figures/compare_grad/01/03.jpg} &
    \includegraphics[trim={1cm 0cm 4cm 0cm}, clip, width=\colwidth\linewidth]{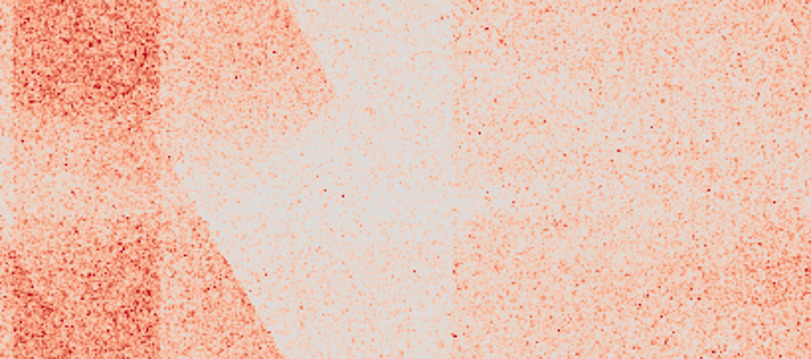} \\
    \includegraphics[width=\colwidth\linewidth]{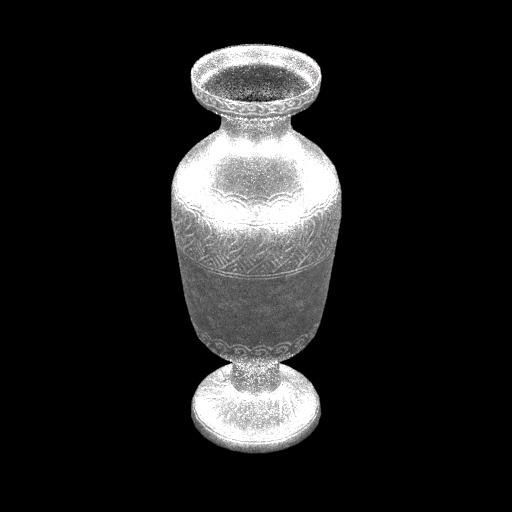} &
    \includegraphics[width=\colwidth\linewidth]{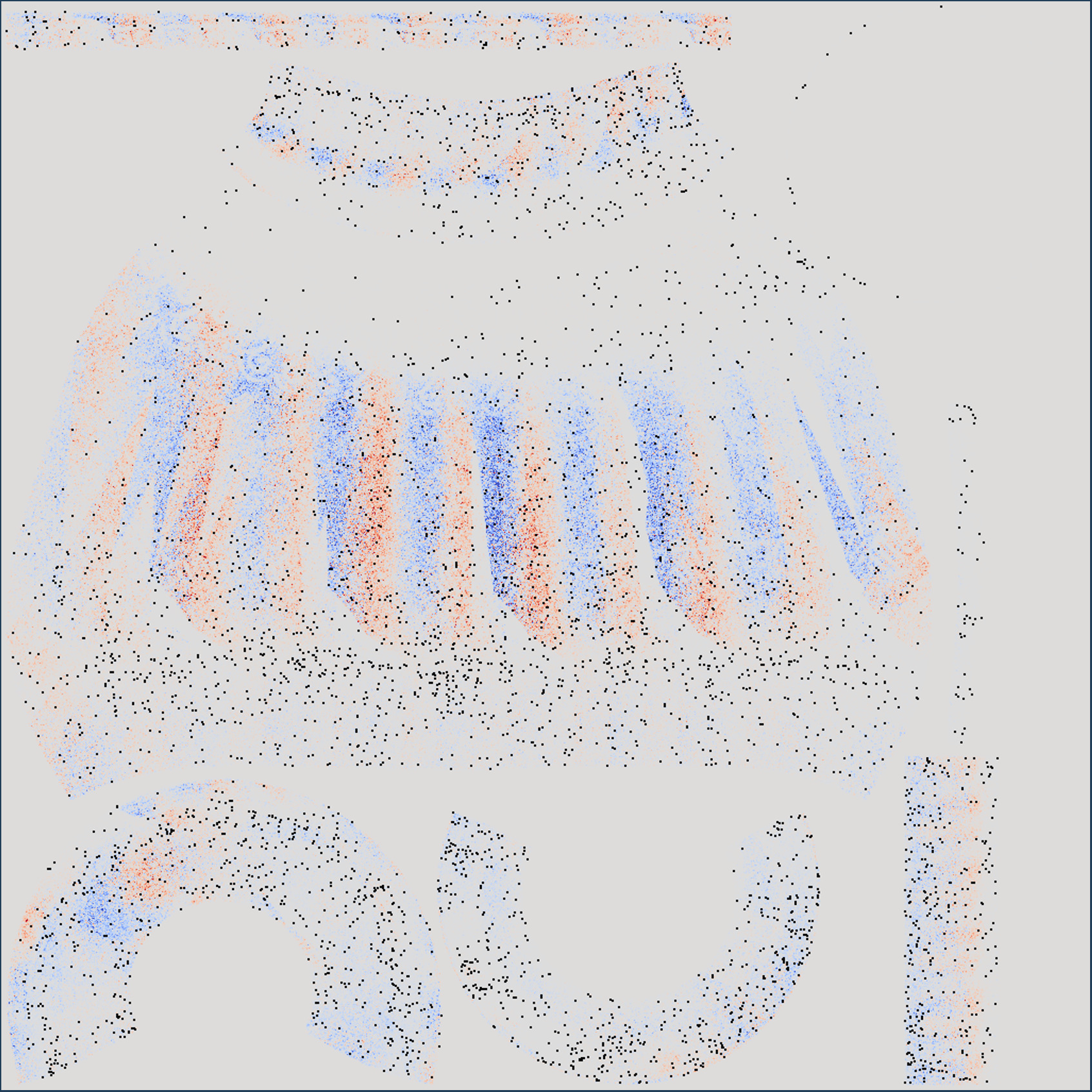} &
    \includegraphics[width=\colwidth\linewidth]{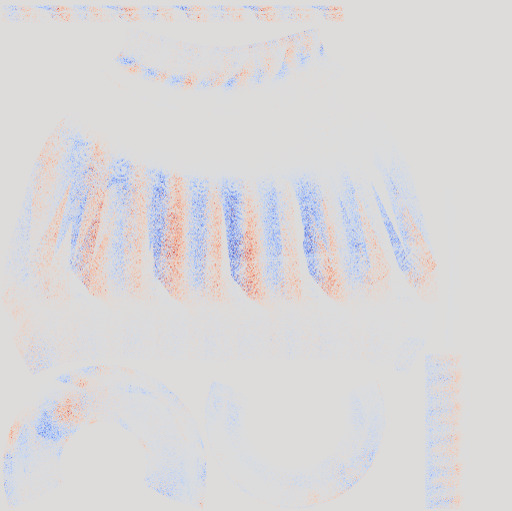} &
    \includegraphics[width=\colwidth\linewidth]{figures/compare_grad/00/03.jpg} &
    \includegraphics[width=\colwidth\linewidth]{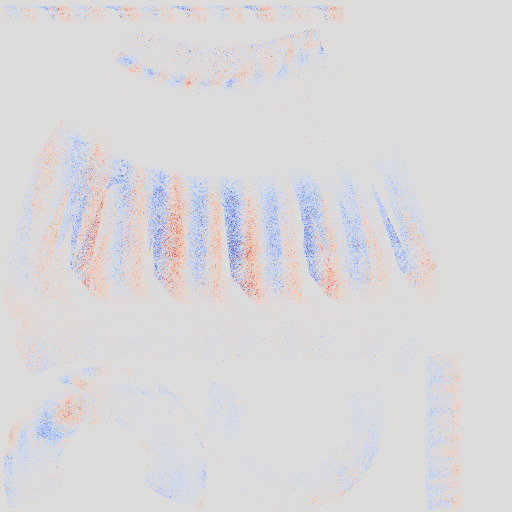} \\
    \includegraphics[width=\colwidth\linewidth]{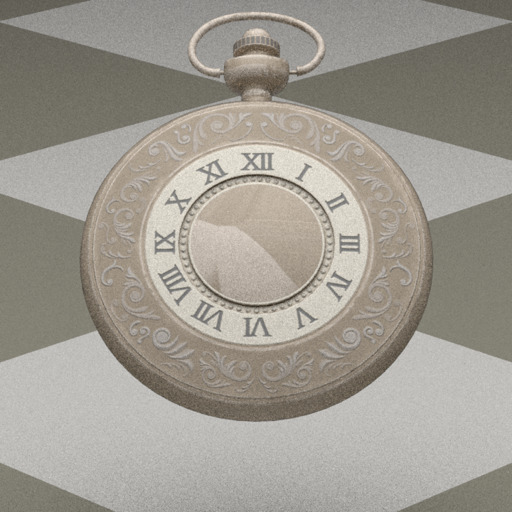} &
    \includegraphics[width=\colwidth\linewidth]{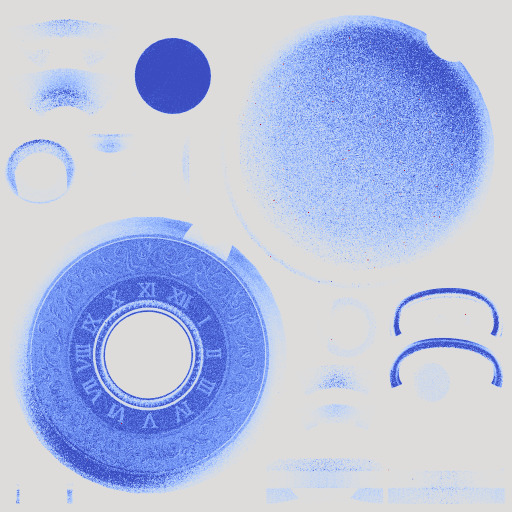} &
    \includegraphics[width=\colwidth\linewidth]{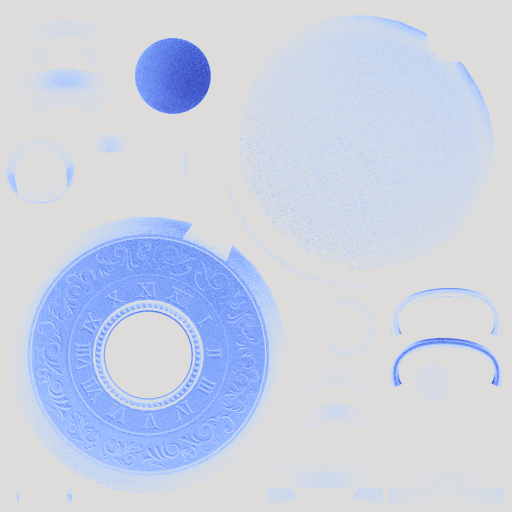} &
    \includegraphics[width=\colwidth\linewidth]{figures/compare_grad/02/03.png} &
    \includegraphics[width=\colwidth\linewidth]{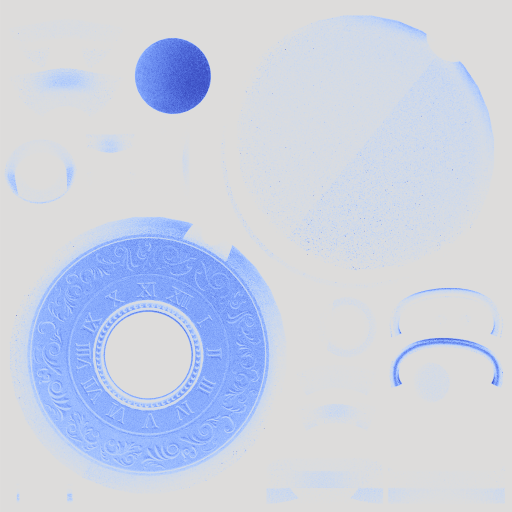} \\
    \includegraphics[width=\colwidth\linewidth]{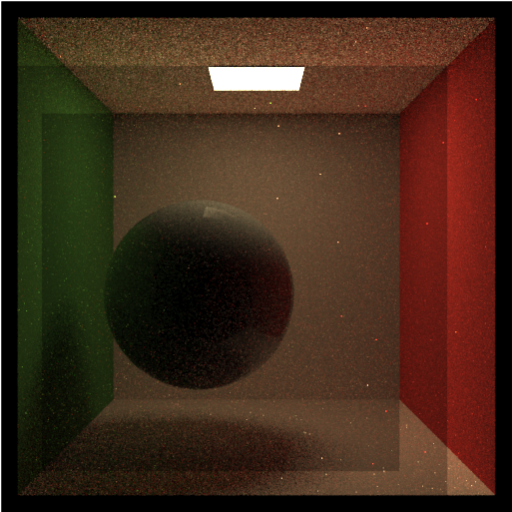} &
    \includegraphics[width=\colwidth\linewidth]{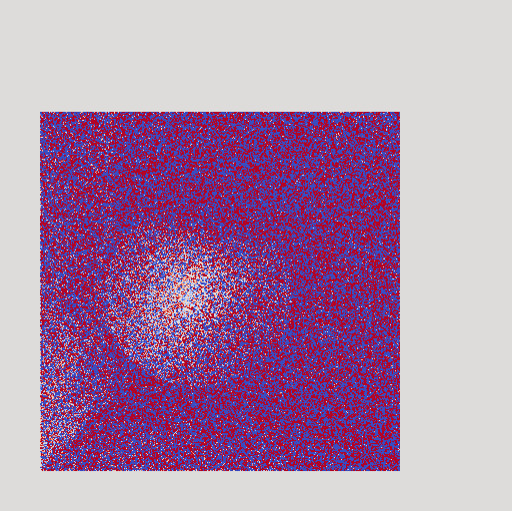} &
    \includegraphics[width=\colwidth\linewidth]{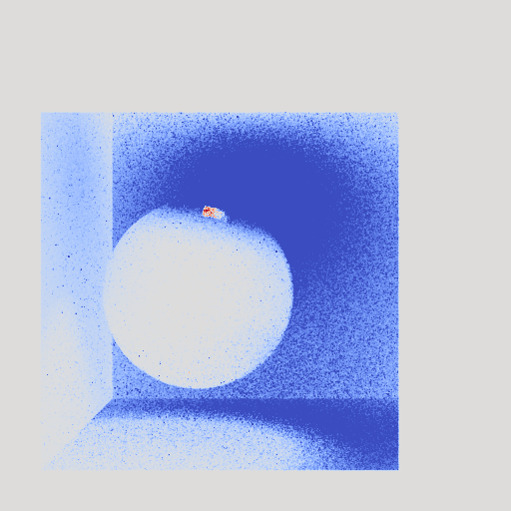} &
    \includegraphics[width=\colwidth\linewidth]{figures/compare_grad/03/03.jpg} &
    \includegraphics[width=\colwidth\linewidth]{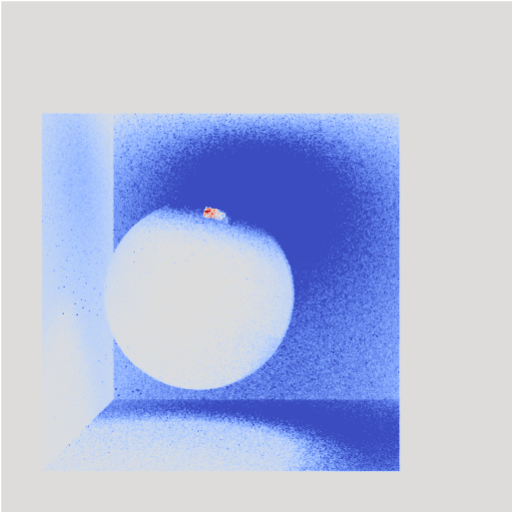} \\
};

\node[col_label] at (grid-1-1.north) {Scene};
\node[col_label] at (grid-1-2.north) {P-PRB};
\node[col_label] at (grid-1-3.north) {P-RB};
\node[col_label] at (grid-1-4.north) {Ours};
\node[col_label] at (grid-1-5.north) {Conv. AD};

\node[row_label_left] at (grid-1-1.west) {Living Room};
\node[row_label_left] at (grid-2-1.west) {Aluminium Vase};
\node[row_label_left] at (grid-3-1.west) {Clock};
\node[row_label_left] at (grid-4-1.west) {Cornell Box};

\node[row_label_right] at (grid-1-5.east) {Diffuse Texture};
\node[row_label_right] at (grid-2-5.east) {Normal Map};
\node[row_label_right] at (grid-3-5.east) {Roughness Texture};
\node[row_label_right] at (grid-4-5.east) {Polarizer $\theta$};

\node[rebox] at (grid-1-2.south east) {RE: 1.4939};
\node[rebox] at (grid-1-3.south east) {RE: 0.3907};
\node[rebox] at (grid-1-4.south east) {RE: 0.3815};
\node[rebox] at (grid-2-2.south east) {RE: 0.6905};
\node[rebox] at (grid-2-3.south east) {RE: 0.1398};
\node[rebox] at (grid-2-4.south east) {RE: 0.1398};
\node[rebox] at (grid-3-2.south east) {RE: 2.7589};
\node[rebox] at (grid-3-3.south east) {RE: 0.1235};
\node[rebox] at (grid-3-4.south east) {RE: 0.1235};
\node[rebox] at (grid-4-2.south east) {RE: 0.3269};
\node[rebox] at (grid-4-3.south east) {RE: 0.0209};
\node[rebox] at (grid-4-4.south east) {RE: 0.0210};

\end{tikzpicture}

\caption{\textbf{Gradient Analysis.} We visualize gradient images and compare against the Conv.\ AD reference. Columns report P-PRB, P-RB, and our method, with each gradient tile annotated by its \emph{relative error (RE)} w.r.t.\ Conv.\ AD (preferred over RMSE to better account for small gradient magnitudes). Rows correspond to different scenes with their associated parameters listed on the right.}
\label{fig:gradient_analysis}
\vspace{-0.5cm}
\end{figure*}

We first validate gradient correctness in \cref{fig:gradient_analysis} by comparing our method against Mitsuba's conventional automatic differentiation (Conv.~AD) and polarized variants of RB and PRB. The figure reports reverse-mode gradients (first three rows) and forward-mode gradients (last row) across multiple scenes and materials.

We use conventional AD as the reference. Since our method and P-RB are unbiased, both produce gradients that closely match the AD baseline. In contrast, the polarized extension of PRB based on noise-regularized matrix inversion (P-PRB) often yields incorrect gradients. The discrepancy becomes more pronounced when incident polarization strongly affects the objective, such as when optimizing surface normals or roughness textures (rows~2 and~3 in \cref{fig:gradient_analysis}).

To further isolate this limitation, we evaluate the gradient in the Cornell Box scene with respect to the linear polarizer rotation parameter~$\theta$. Here, the sampled BSDF contribution is a rank-deficient Mueller matrix (linear polarizer). The baseline P-PRB relies on inverting the sampled BSDF's Mueller matrix, which is not well-defined. As a result, P-PRB breaks down in this setting, making it ineffective for differentiating through a linear polarizer, which is a canonical optical element in polarimetric scenes.

\subsection{Impact of Polarization on Inverse Rendering}
\label{subsec:results_application}
\begin{figure*}[t]
\centering

\newcommand{\mywidth}{0.18}

\begin{tikzpicture}[
    every node/.style={inner sep=0, outer sep=0},
    col_label/.style={font=\sffamily\bfseries\small, anchor=south, yshift=1mm},
    row_label_left/.style={
      font=\sffamily\bfseries\tiny, 
      rotate=90,
      anchor=south,
      yshift=1mm,
      text width=18mm,     
      align=center
    },
    row_label_right/.style={
      font=\sffamily\bfseries\tiny, 
      rotate=270,
      anchor=south,
      yshift=1mm,
      text width=18mm,     
      align=center
    },
]

\matrix (mymatrix) [
    matrix of nodes,
    nodes={anchor=center},
    column sep=0.8mm,
    row sep=0.8mm
] {
    \includegraphics[width=\mywidth\linewidth]{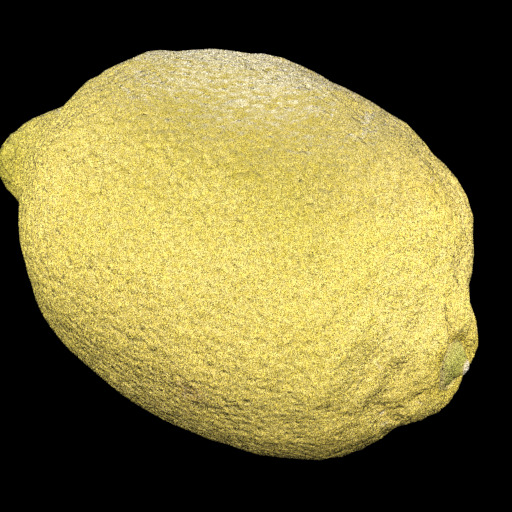}   &
    \includegraphics[width=\mywidth\linewidth]{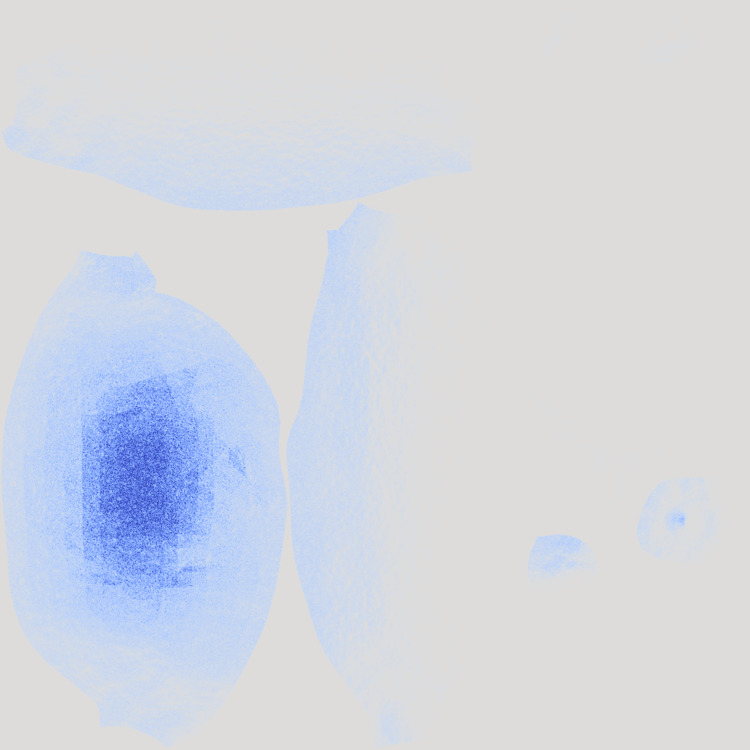}   &
    \includegraphics[width=\mywidth\linewidth]{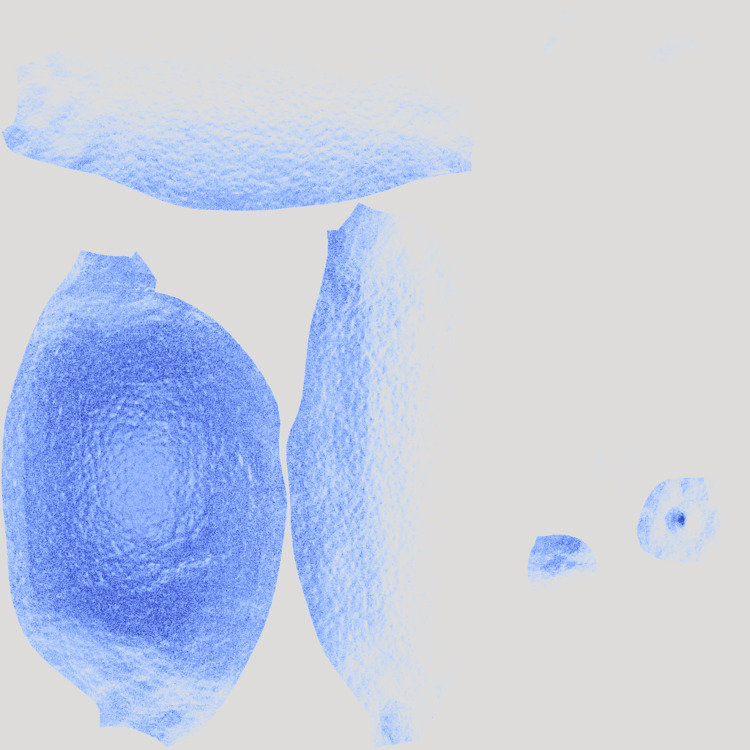}   &
    \includegraphics[width=\mywidth\linewidth]{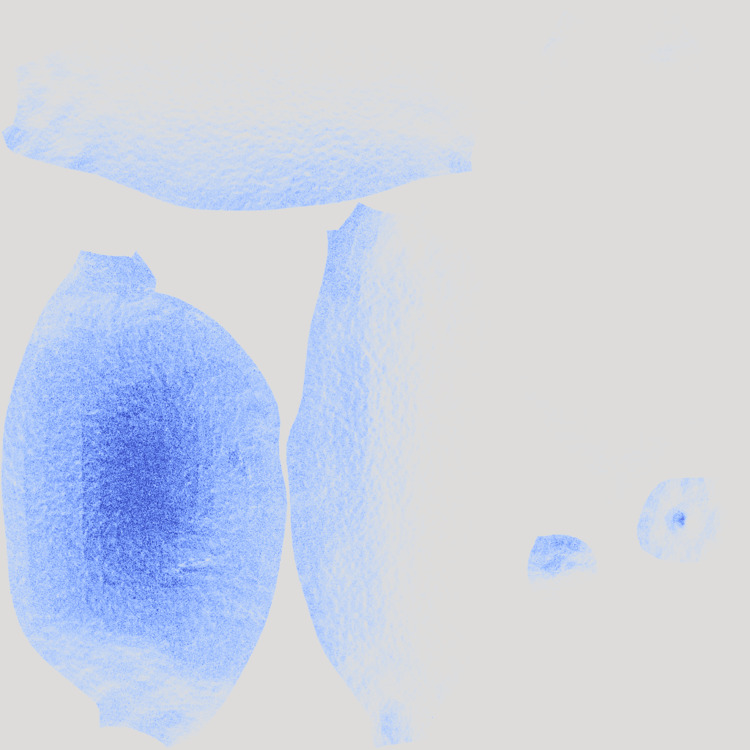}   &
    \includegraphics[width=\mywidth\linewidth]{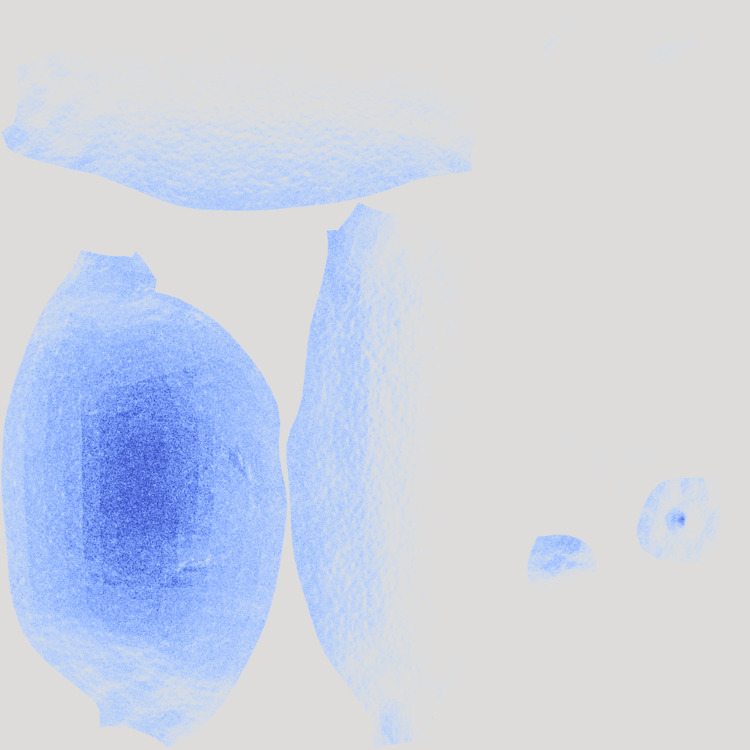}  \\
    \includegraphics[width=\mywidth\linewidth]{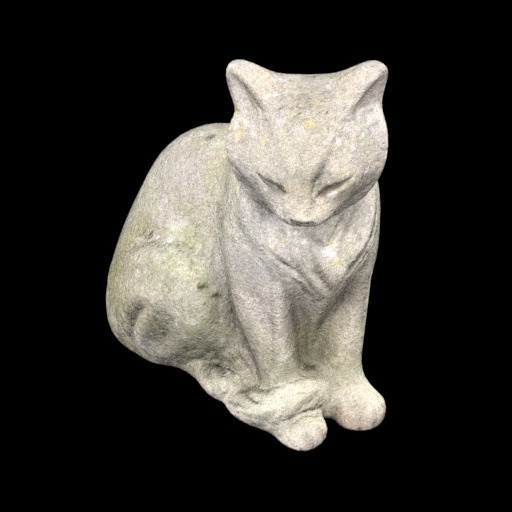}   &
    \includegraphics[width=\mywidth\linewidth]{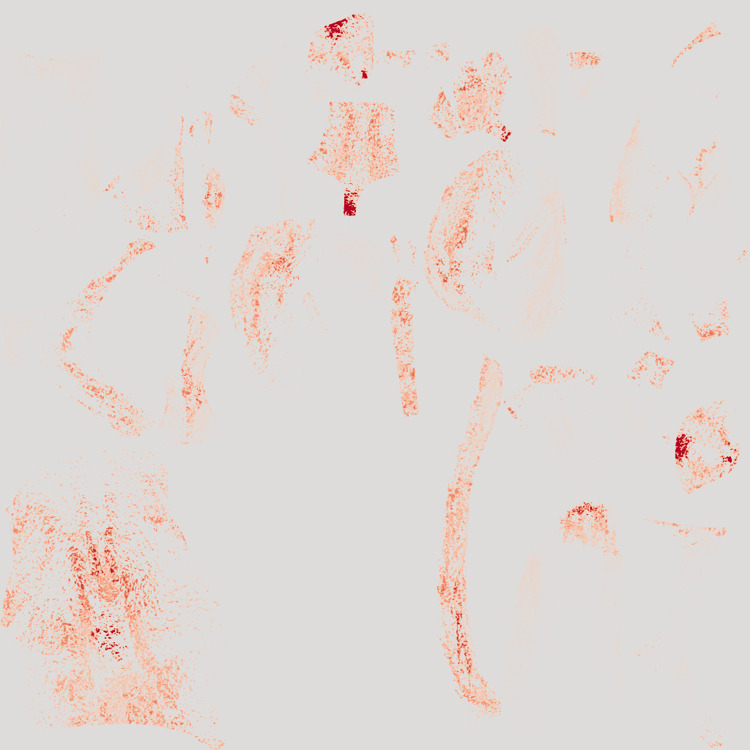}   &
    \includegraphics[width=\mywidth\linewidth]{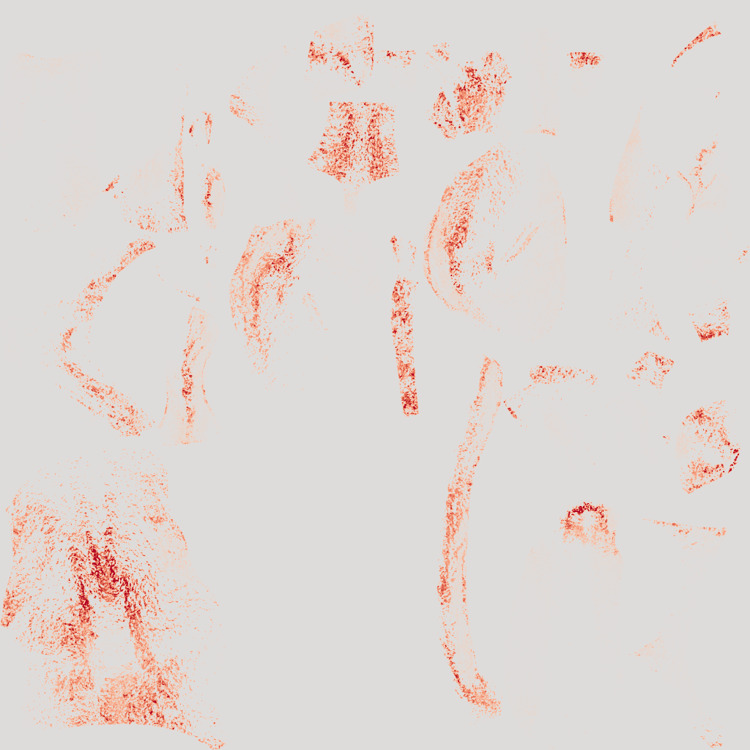}   &
    \includegraphics[width=\mywidth\linewidth]{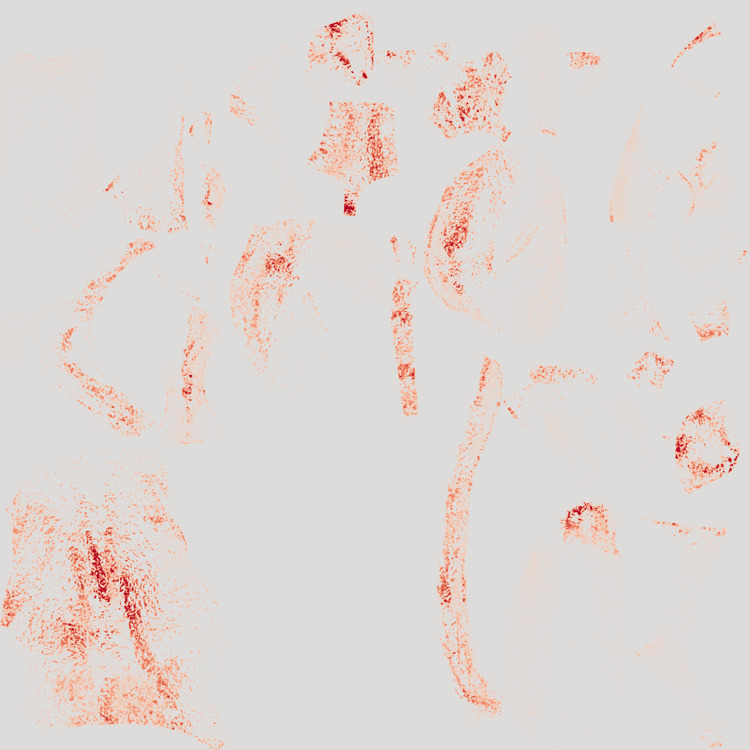}   &
    \includegraphics[width=\mywidth\linewidth]{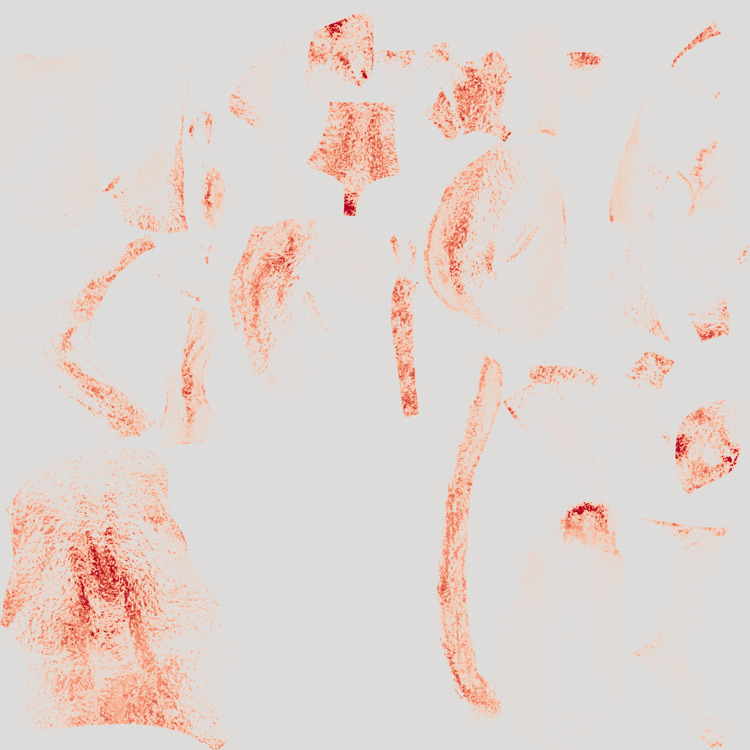}  \\
    \includegraphics[width=\mywidth\linewidth]{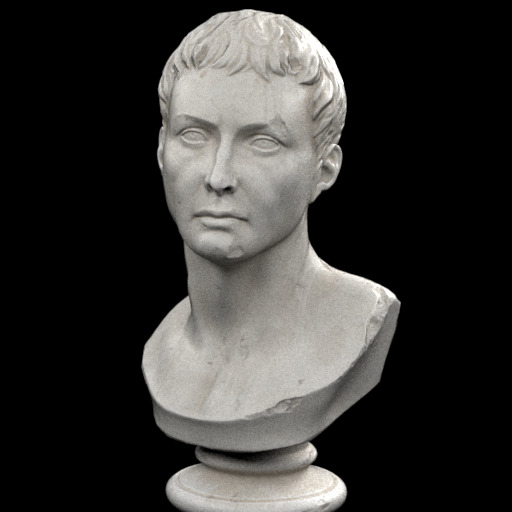}   &
    \includegraphics[width=\mywidth\linewidth]{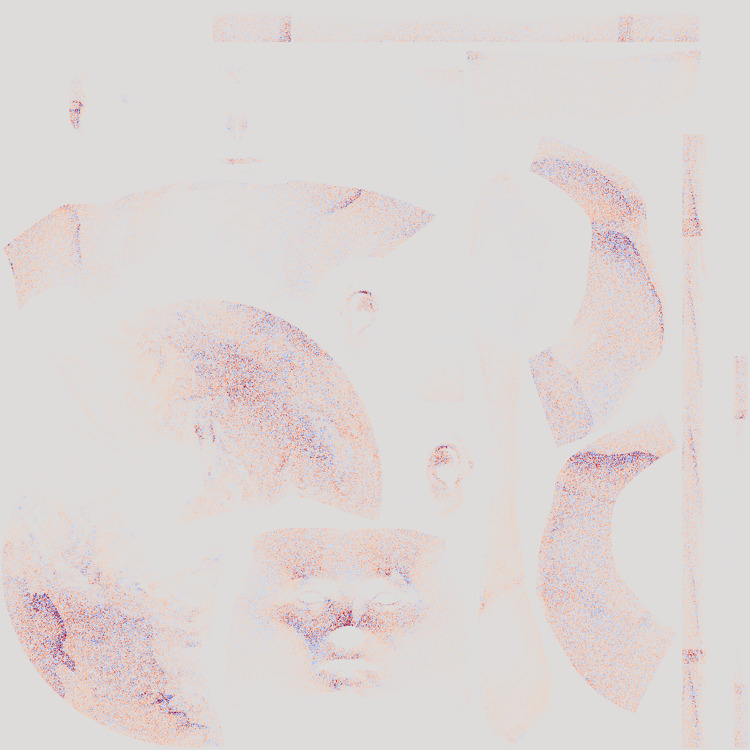}   &
    \includegraphics[width=\mywidth\linewidth]{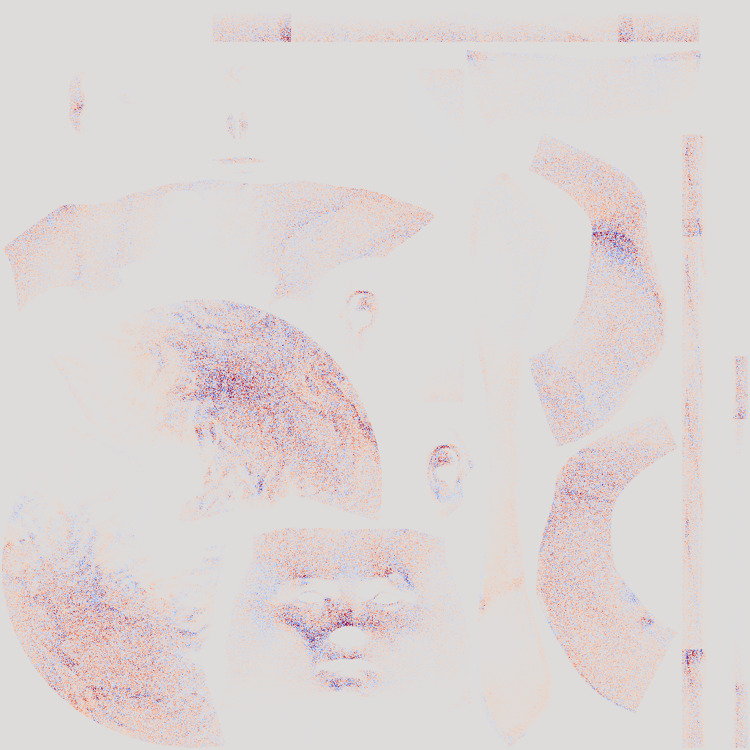}   &
    \includegraphics[width=\mywidth\linewidth]{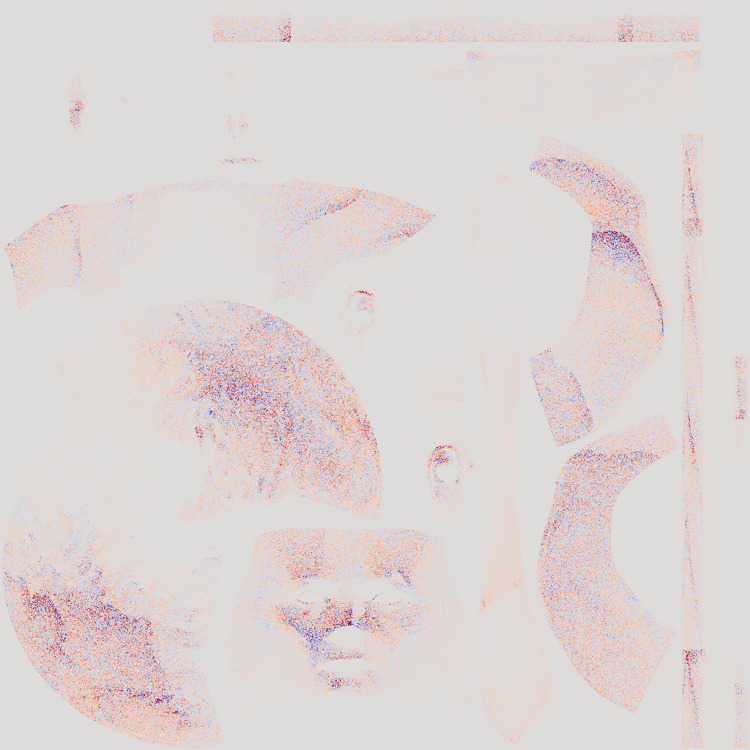}   &
    \includegraphics[width=\mywidth\linewidth]{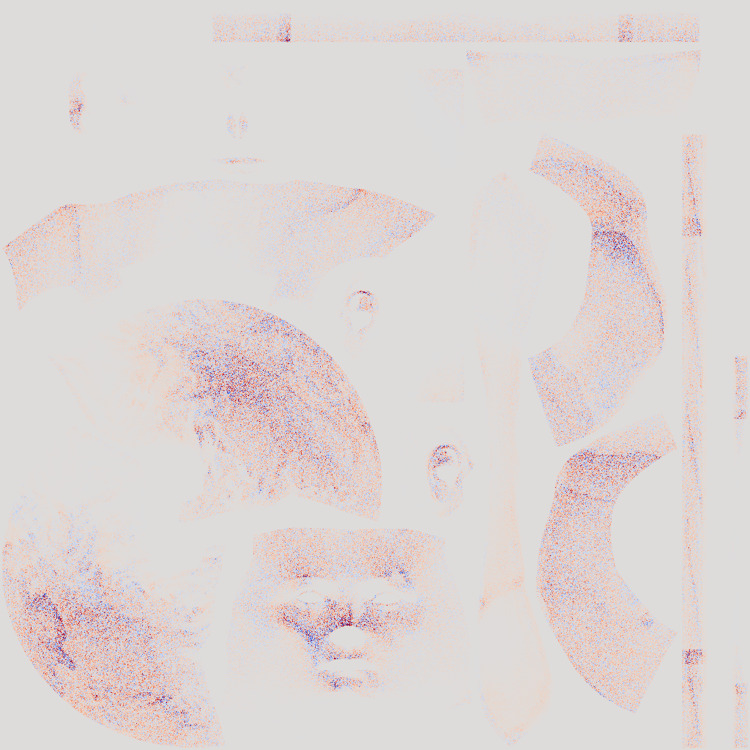}  \\
};

\node[col_label] at (mymatrix-1-1.north) {Scene};
\node[col_label] at (mymatrix-1-2.north) {0};
\node[col_label] at (mymatrix-1-3.north) {45};
\node[col_label] at (mymatrix-1-4.north) {90};
\node[col_label] at (mymatrix-1-5.north) {Left-circular};

\node[row_label_left]  at (mymatrix-1-1.west) {Lemon};
\node[row_label_left]  at (mymatrix-2-1.west) {Cat Statue};
\node[row_label_left]  at (mymatrix-3-1.west) {Marble Bust};

\node[row_label_right] at (mymatrix-1-5.east) {Roughness Texture};
\node[row_label_right] at (mymatrix-2-5.east) {Specular Map};
\node[row_label_right] at (mymatrix-3-5.east) {Normal Map};

\end{tikzpicture}

\caption{Qualitative comparison of reverse-mode gradients under different polarization configurations. We render a virtual light stage setup with right-circular illumination and accumulate gradients over a multi-view camera setup. Columns correspond to polarizer configurations ($0^\circ$, $45^\circ$, $90^\circ$, and left-circular), and rows show gradients with respect to different scene parameters (surface normals, roughness, and specular reflectance). The resulting gradients exhibit distinct spatial structures across polarization states, illustrating that polarization provides complementary information when optimizing different parameters.}
\label{fig:scene_angle_lc_grid}
\vspace{-0.5cm}
\end{figure*}

While the previous section establishes gradient correctness, we now examine the role of polarization in inverse rendering tasks. We first demonstrate the variation of gradients under different polarization configurations. Inspired by the multi-view capture setups of \cite{circular_ghosh2010, zhou2026olatverse, prao20253dpr}, we construct a virtual light stage with right-circular illumination and a centrally placed object observed from multiple views. The object is rendered under four polarization configurations: $0^\circ$, $45^\circ$, $90^\circ$ and left-circular.

\begin{figure*}[t]
\centering
\newcommand{\mywidth}{0.2} 

\begin{tikzpicture}[
    every node/.style={inner sep=0,outer sep=0},
    col_label/.style={font=\sffamily\bfseries\small, anchor=south, yshift=1mm},
    row_label_left/.style={
      font=\sffamily\bfseries\tiny, 
      rotate=90,
      anchor=south,
      yshift=1mm,
      text width=18mm,     
      align=center
    },
    row_label_right/.style={
      font=\sffamily\bfseries\tiny, 
      rotate=270,
      anchor=south,
      yshift=1mm,
      text width=18mm,     
      align=center
    },
    rmse/.style={
        font=\sffamily\bfseries\tiny,
        text=white,
        anchor=south east,
        xshift=-0.8mm,
        yshift=0.6mm,
        fill=black,
        fill opacity=0.45,
        text opacity=1,
        rounded corners=0.6pt,
        inner sep=1.2pt
    }
]

\matrix (mymatrix) [
    matrix of nodes, 
    nodes={anchor=center}, 
    column sep=0.0mm, 
    row sep=0.1mm
] {
    \includegraphics[width=\mywidth\linewidth]{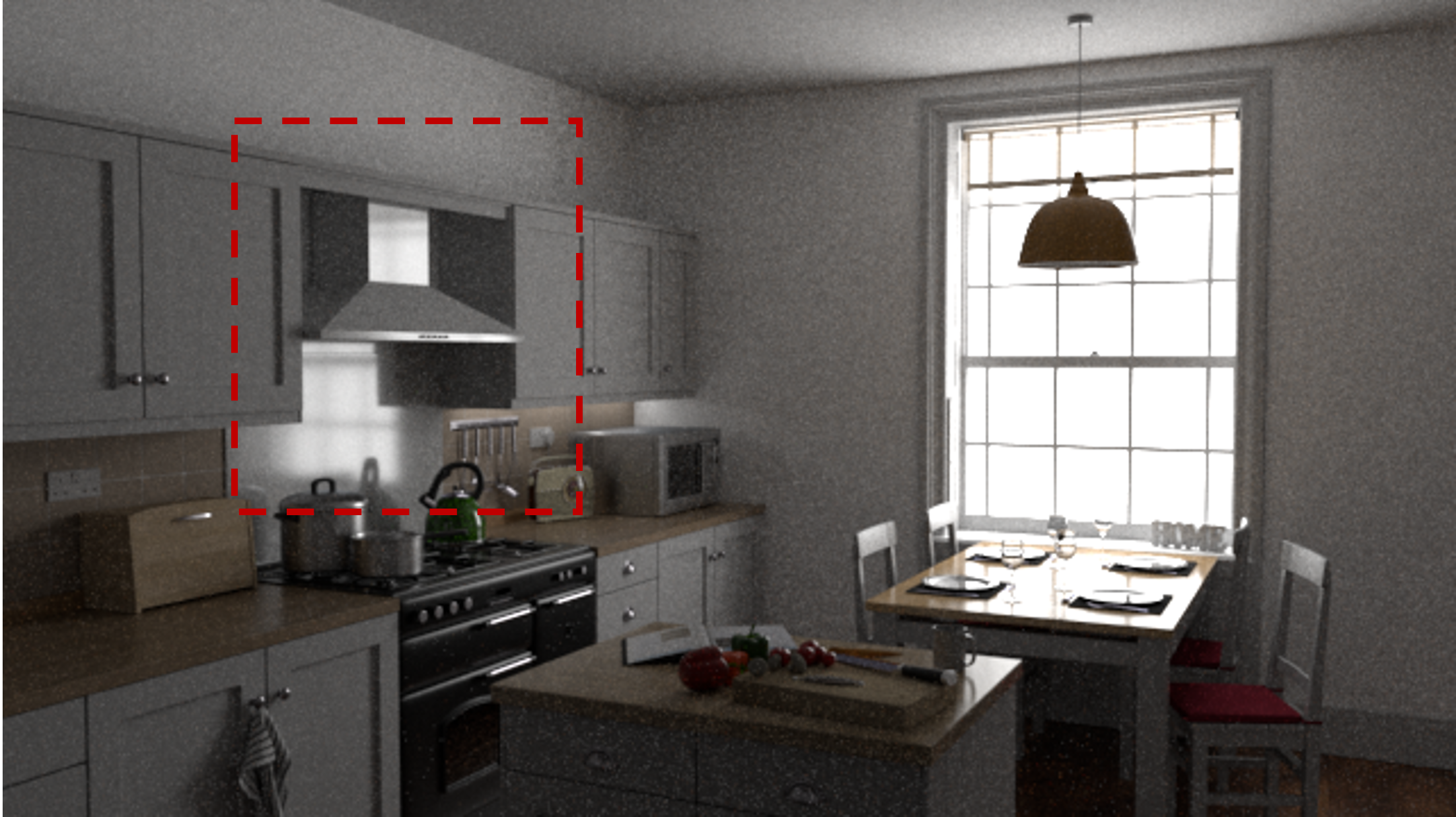} &
    \includegraphics[width=\mywidth\linewidth]{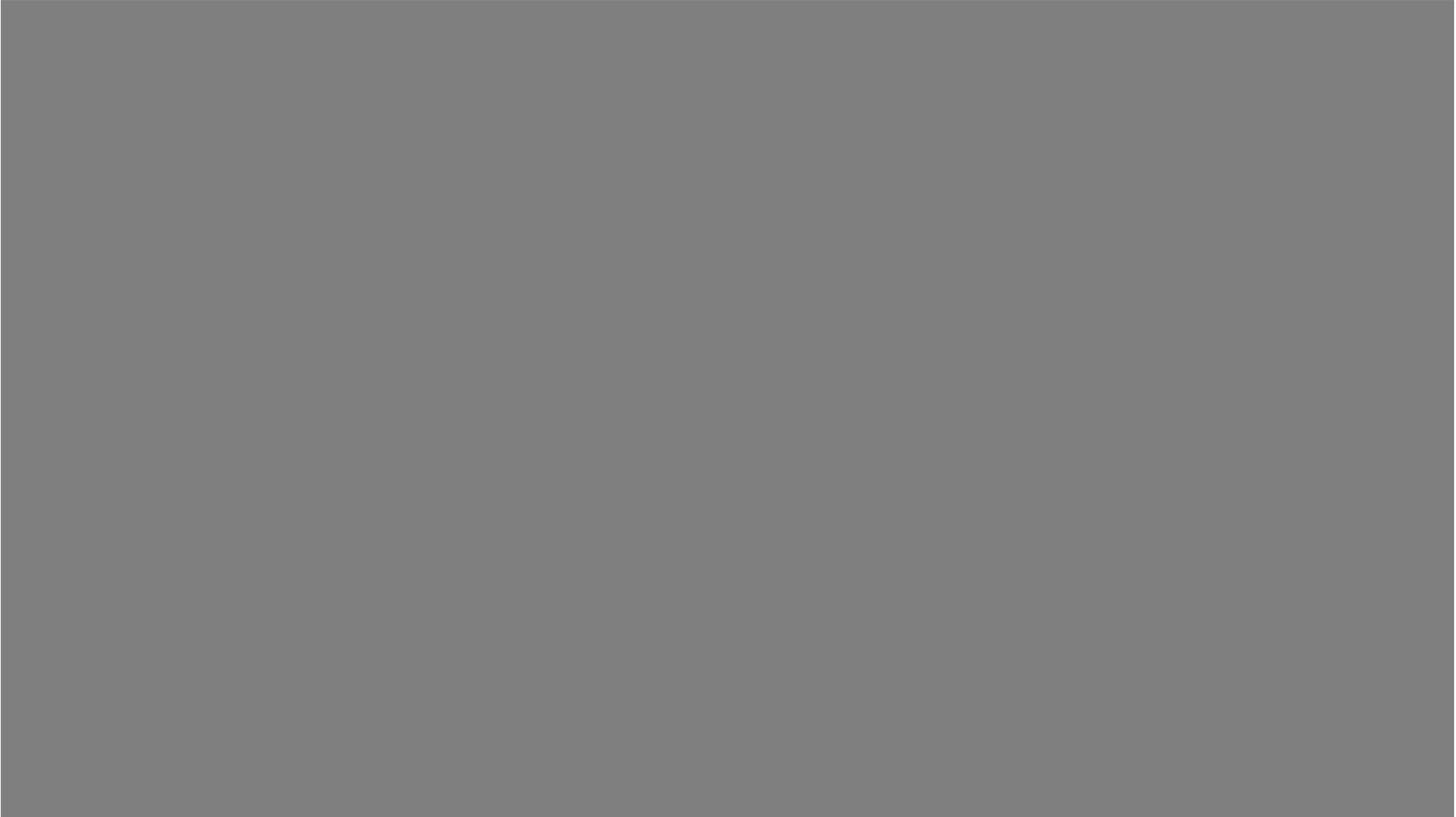} &
    \includegraphics[width=\mywidth\linewidth]{figures/baselines/00/00.png} &
    \includegraphics[width=\mywidth\linewidth]{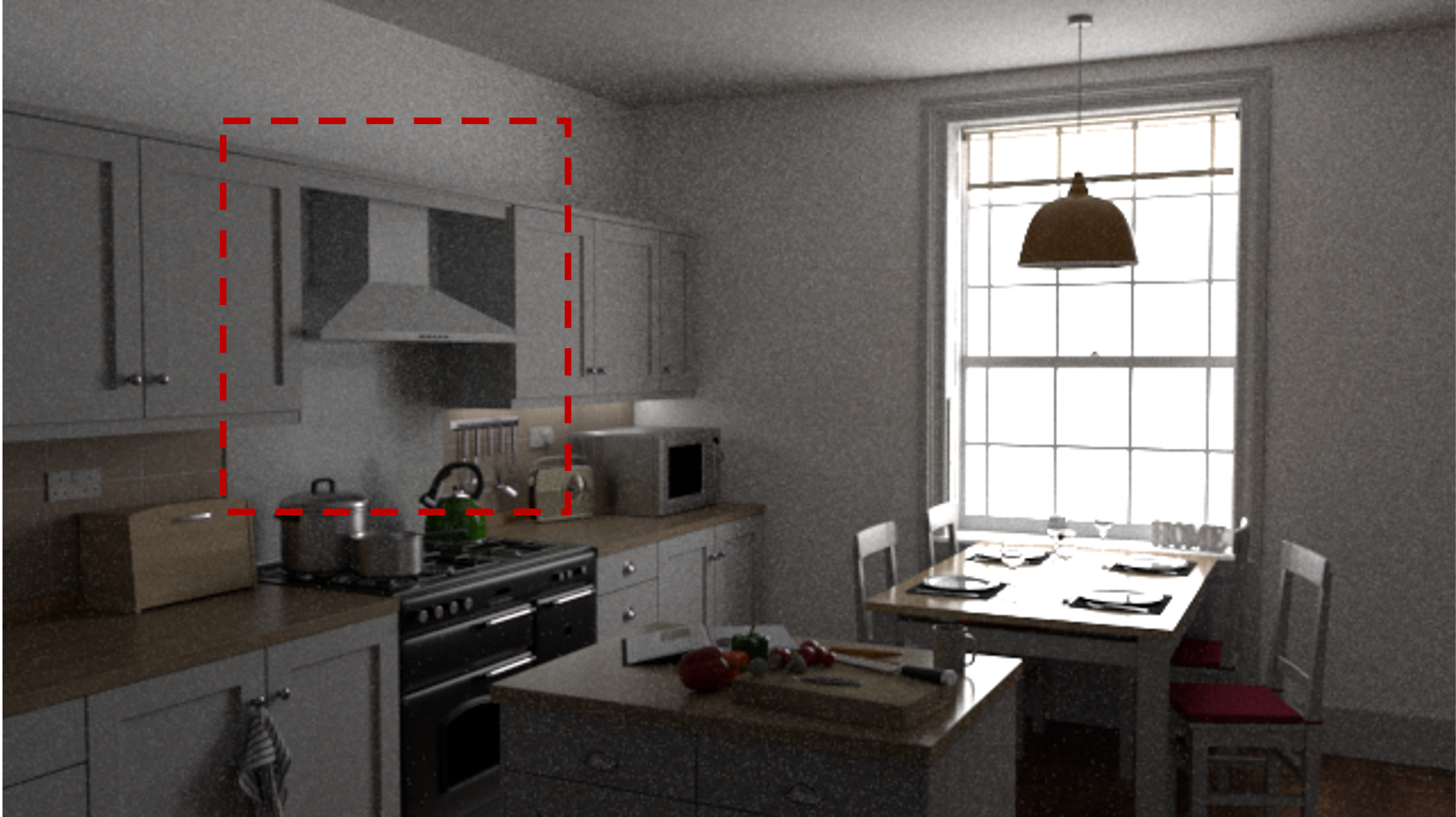} &
    \includegraphics[width=\mywidth\linewidth]{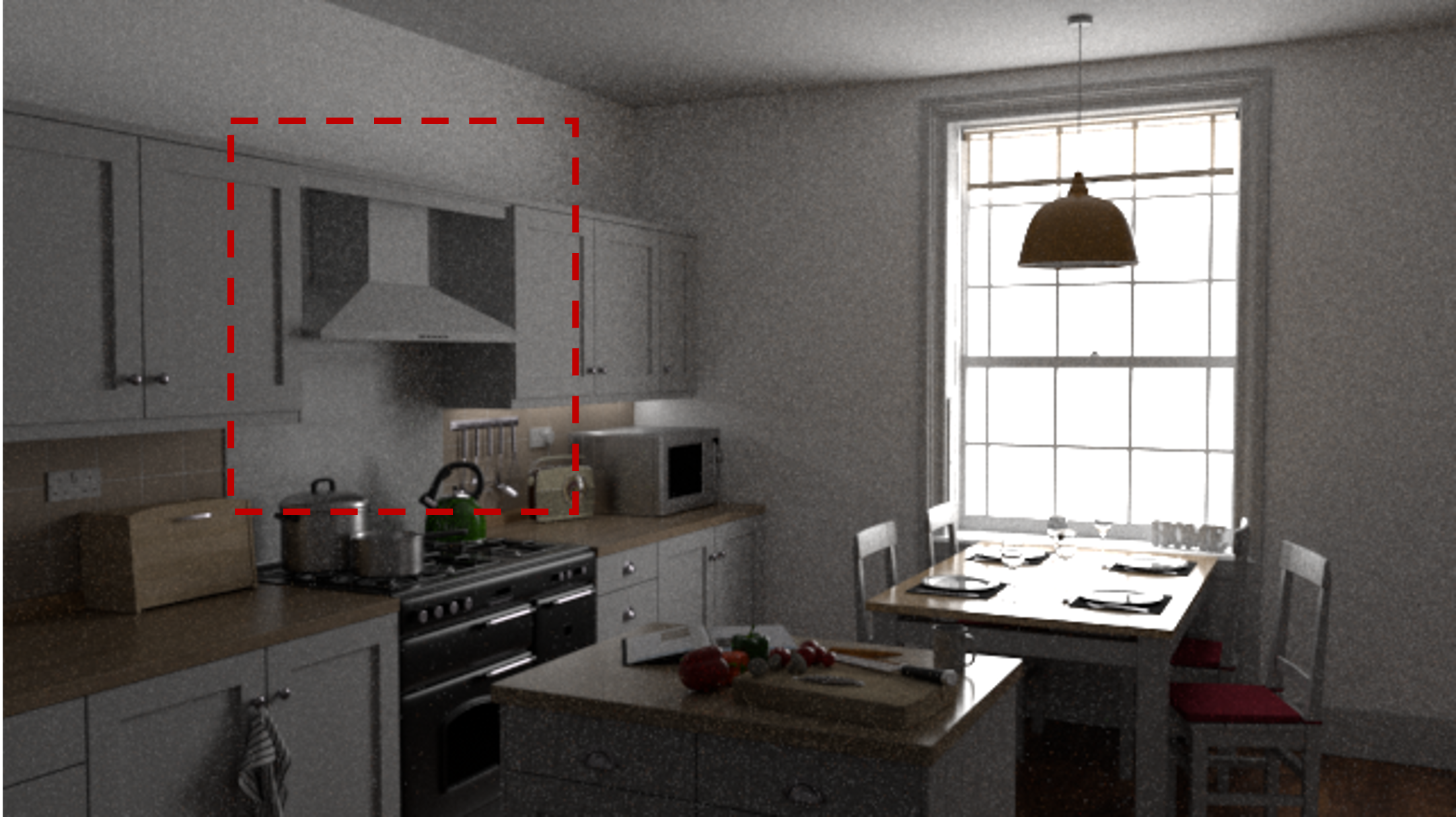} \\
    
    \includegraphics[width=\mywidth\linewidth]{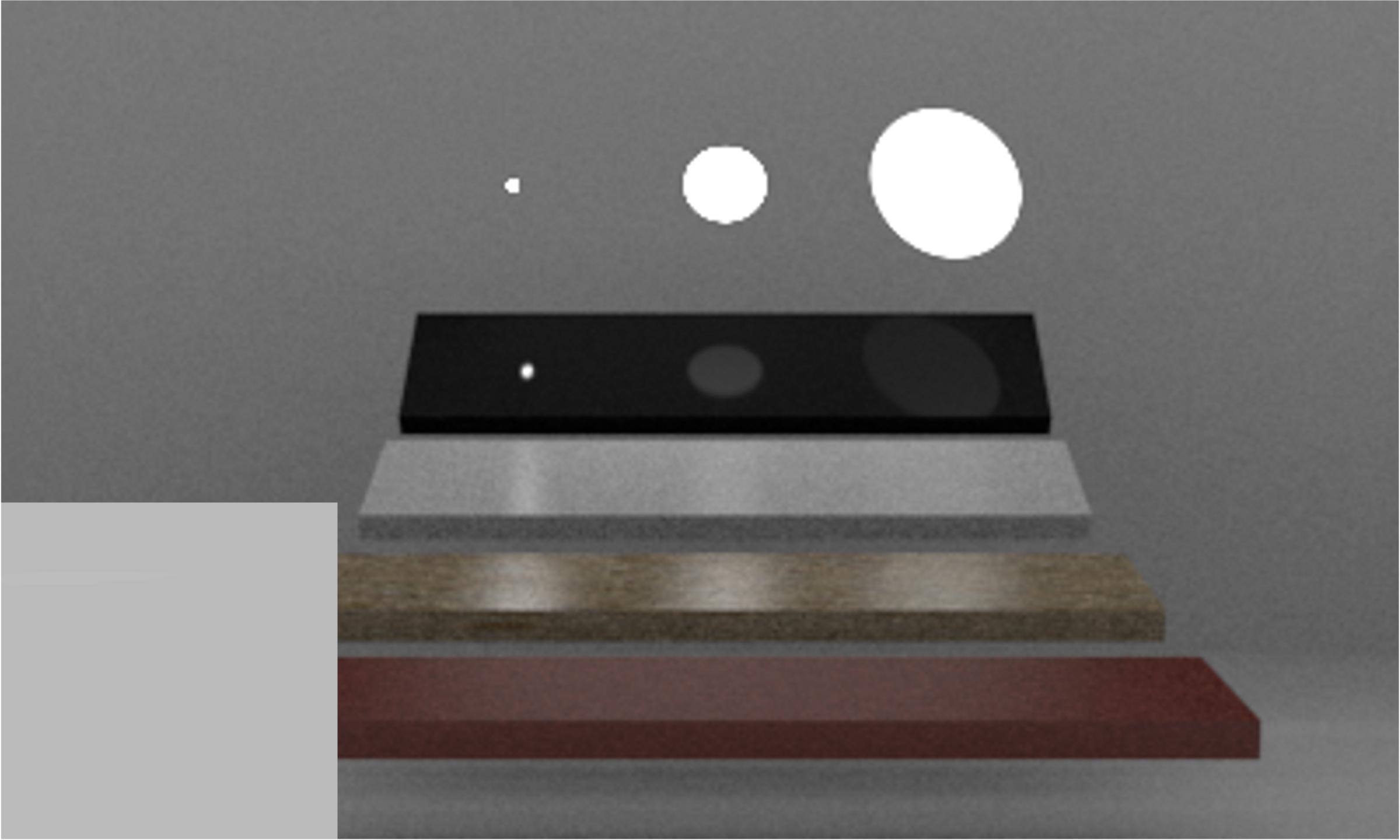} &
    \includegraphics[width=\mywidth\linewidth]{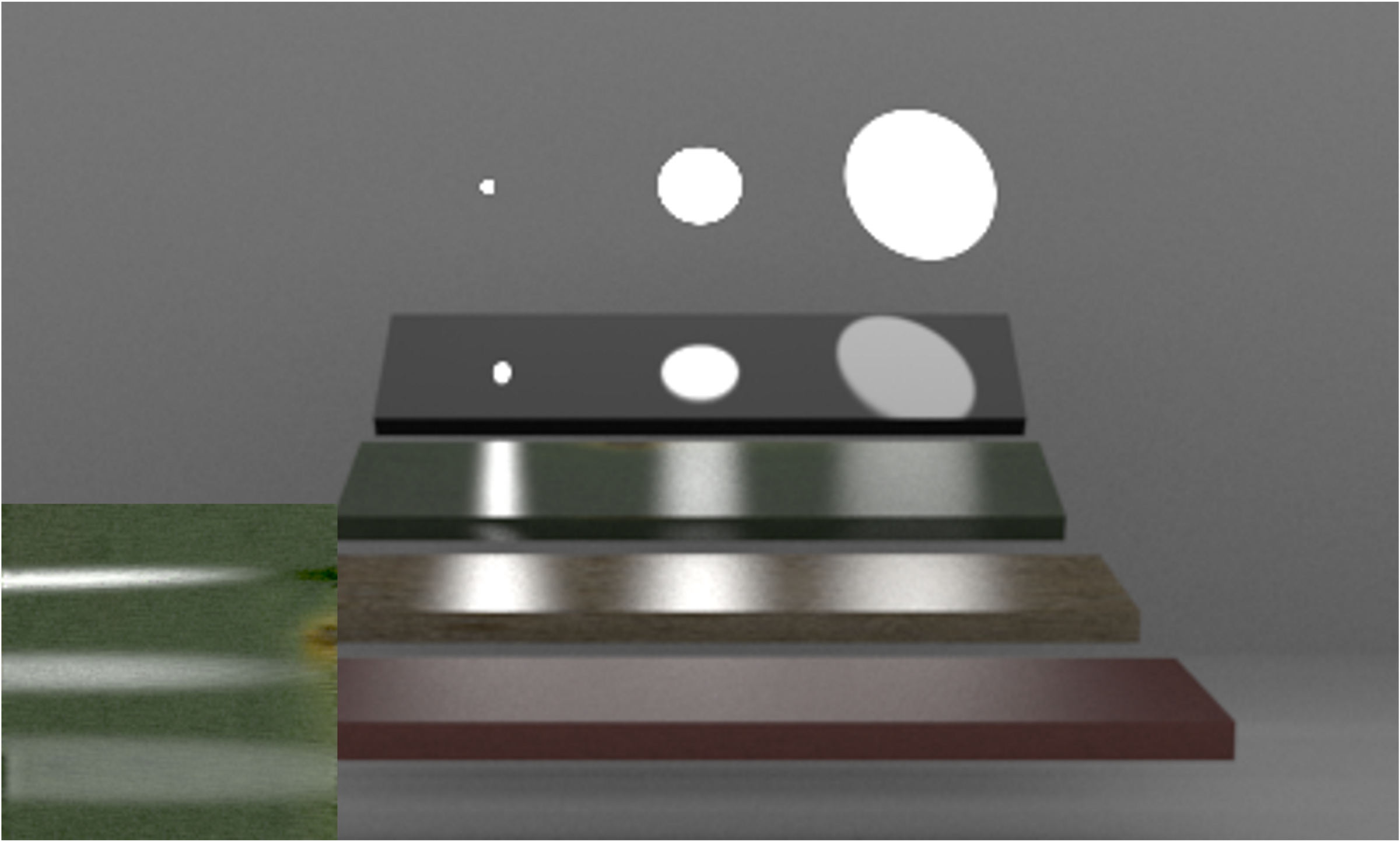} &
    \includegraphics[width=\mywidth\linewidth]{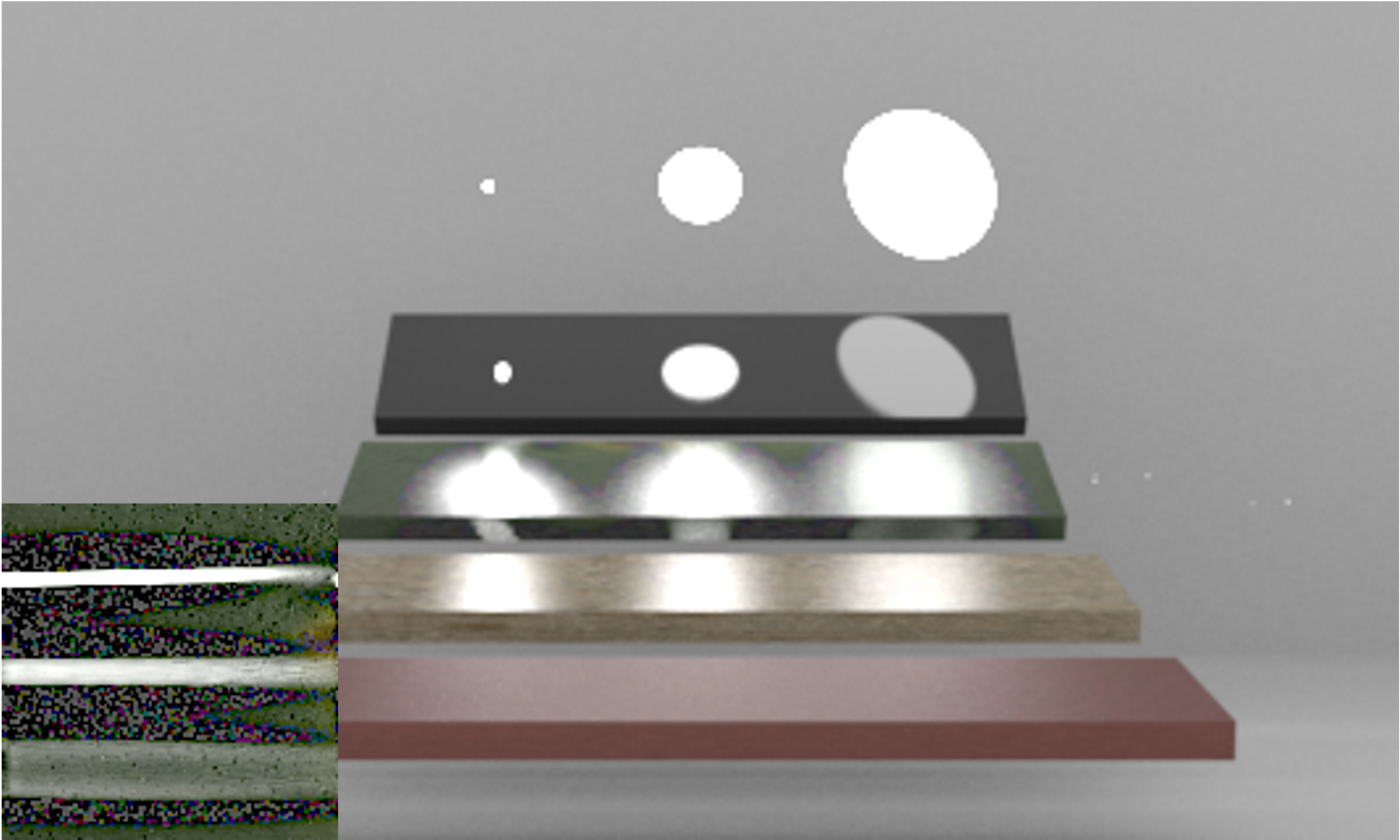} &
    \includegraphics[width=\mywidth\linewidth]{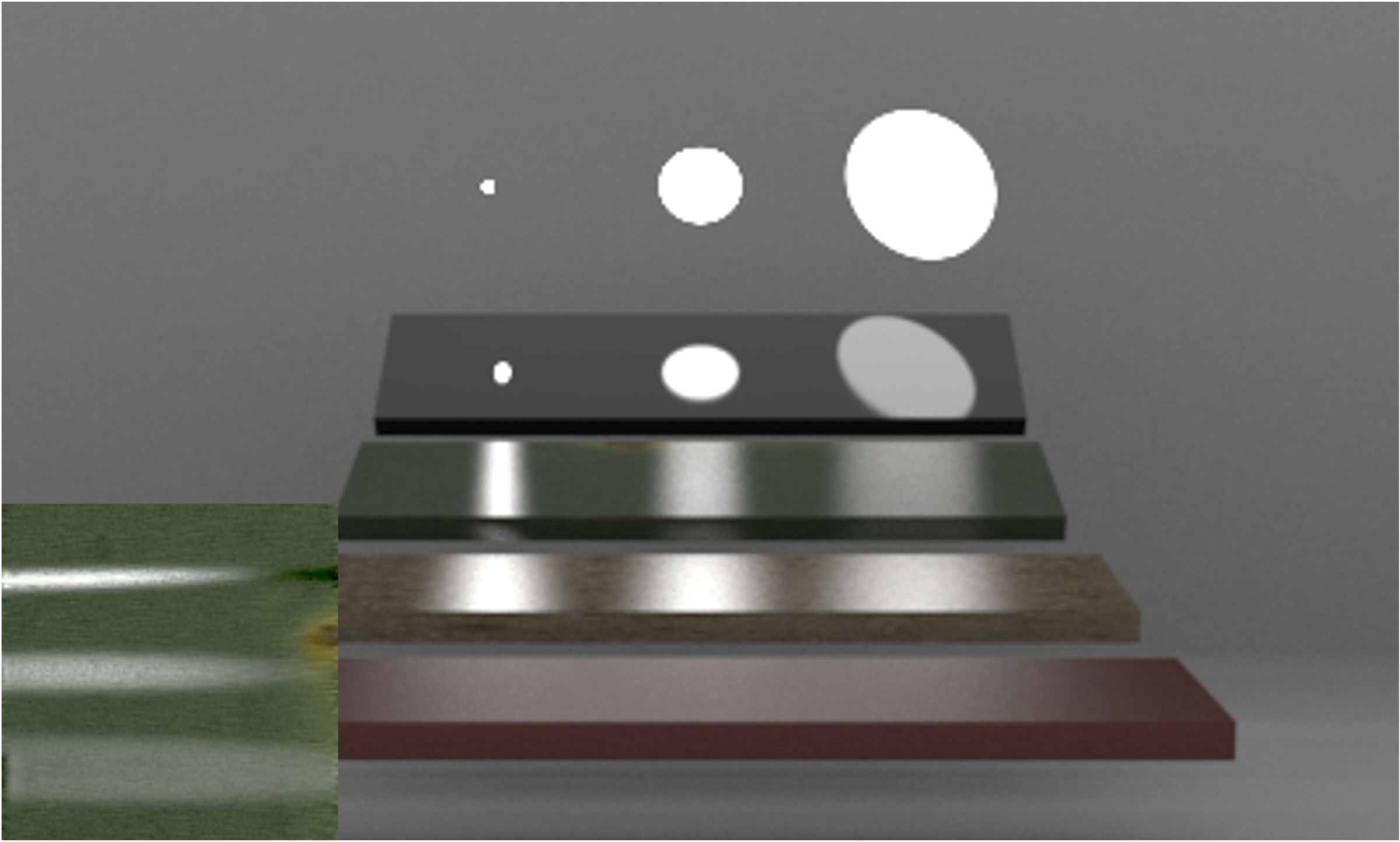} &
    \includegraphics[width=\mywidth\linewidth]{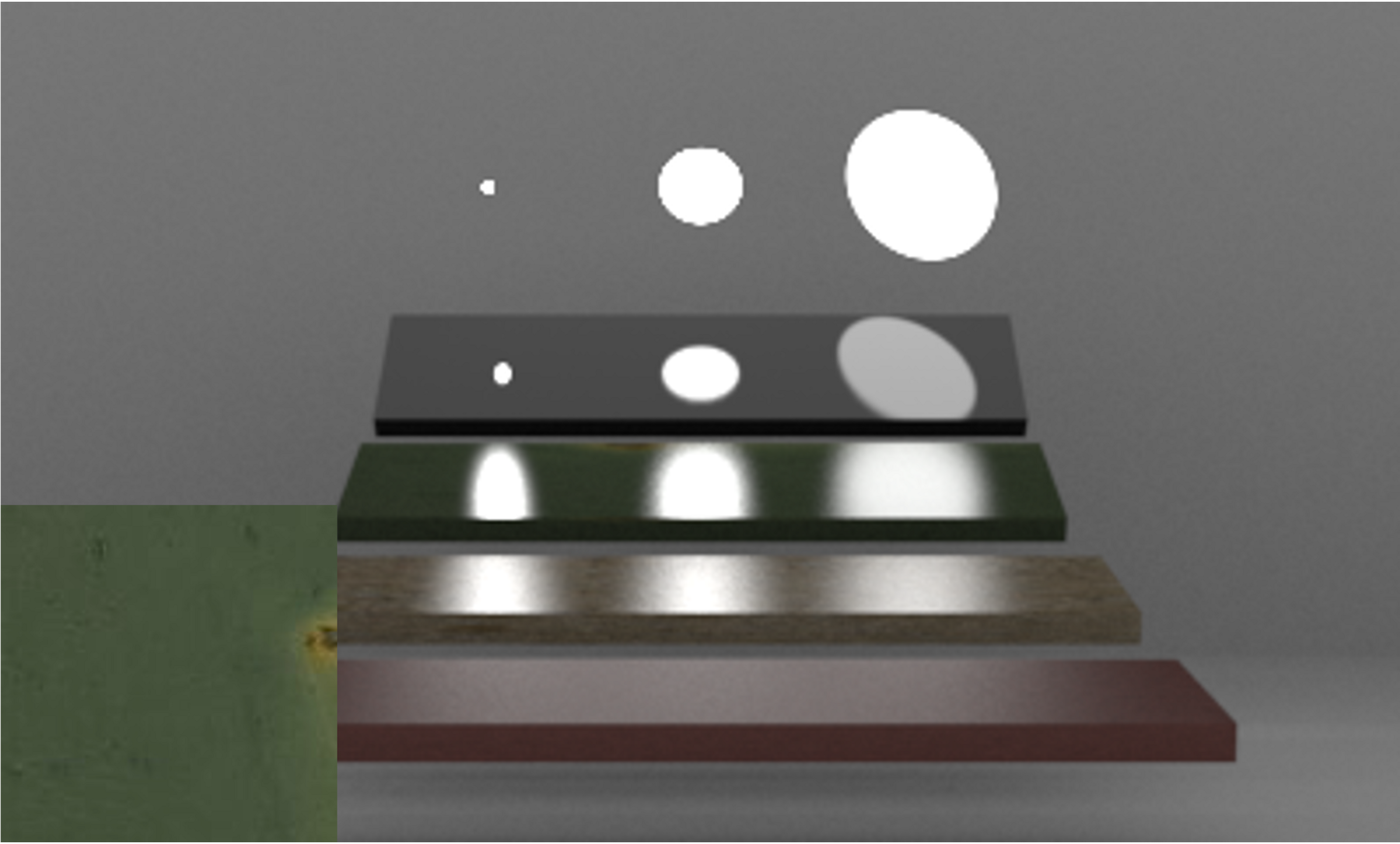} \\
    \includegraphics[width=\mywidth\linewidth]{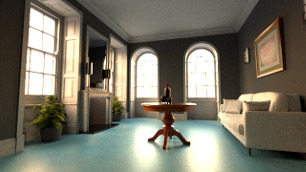} &
    \includegraphics[width=\mywidth\linewidth]{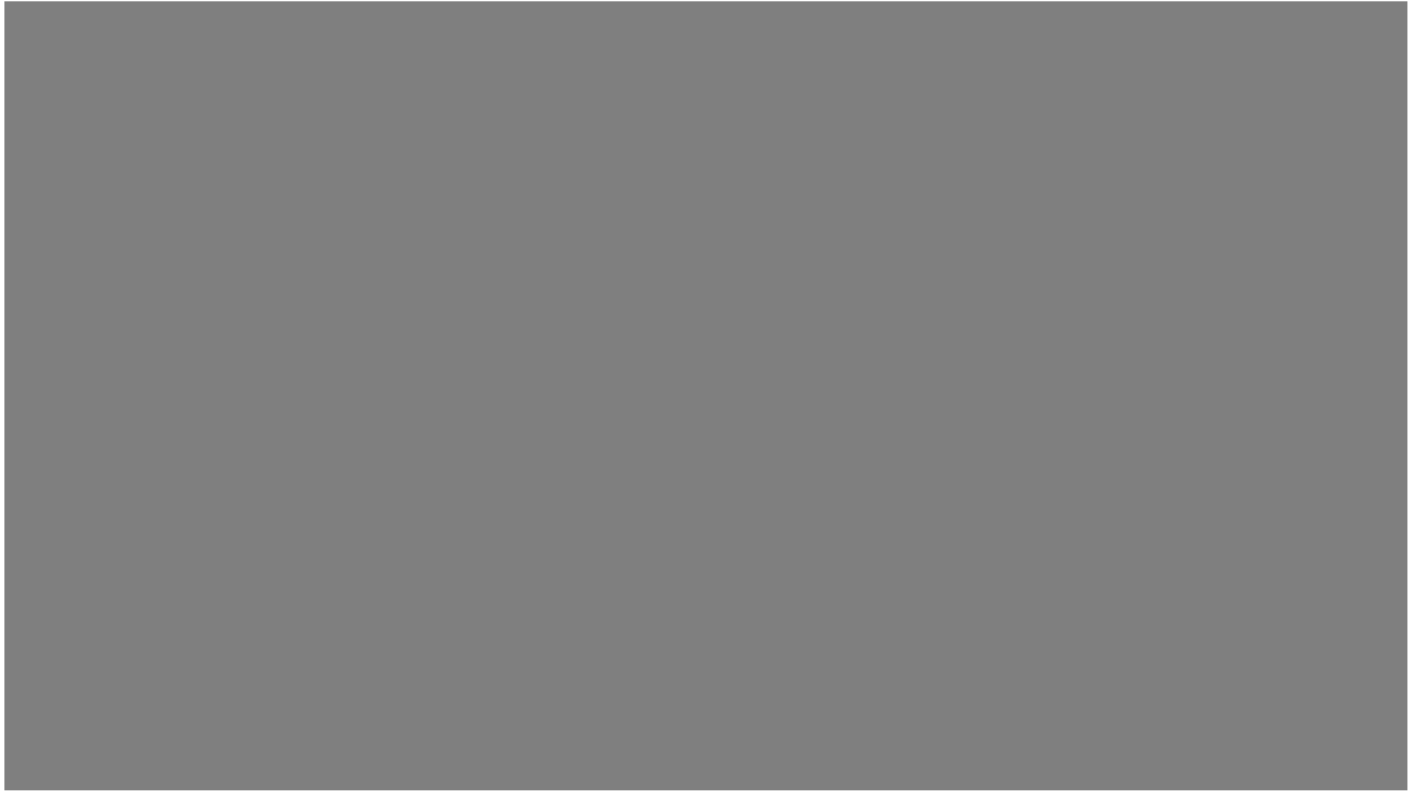} &
    \includegraphics[width=\mywidth\linewidth]{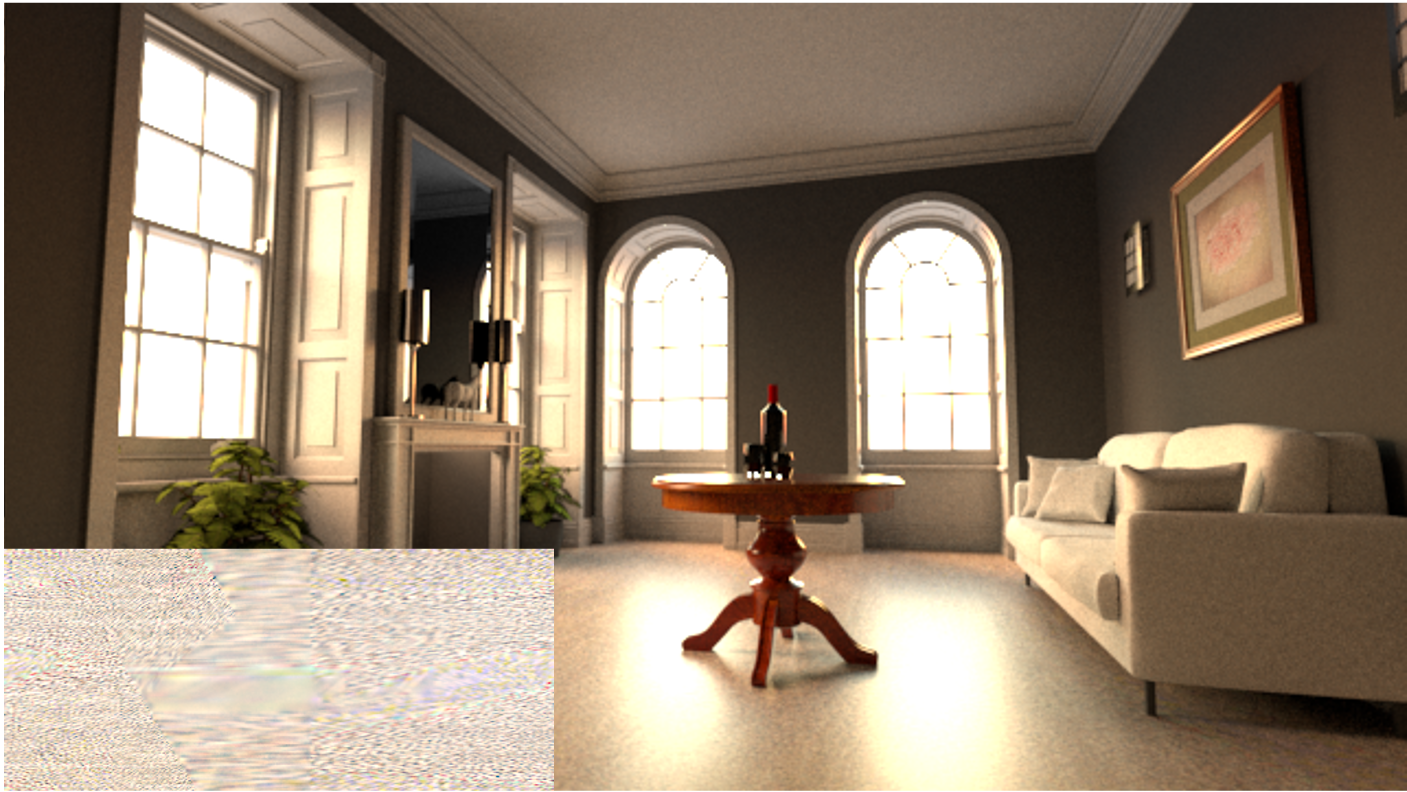} &
    \includegraphics[width=\mywidth\linewidth]{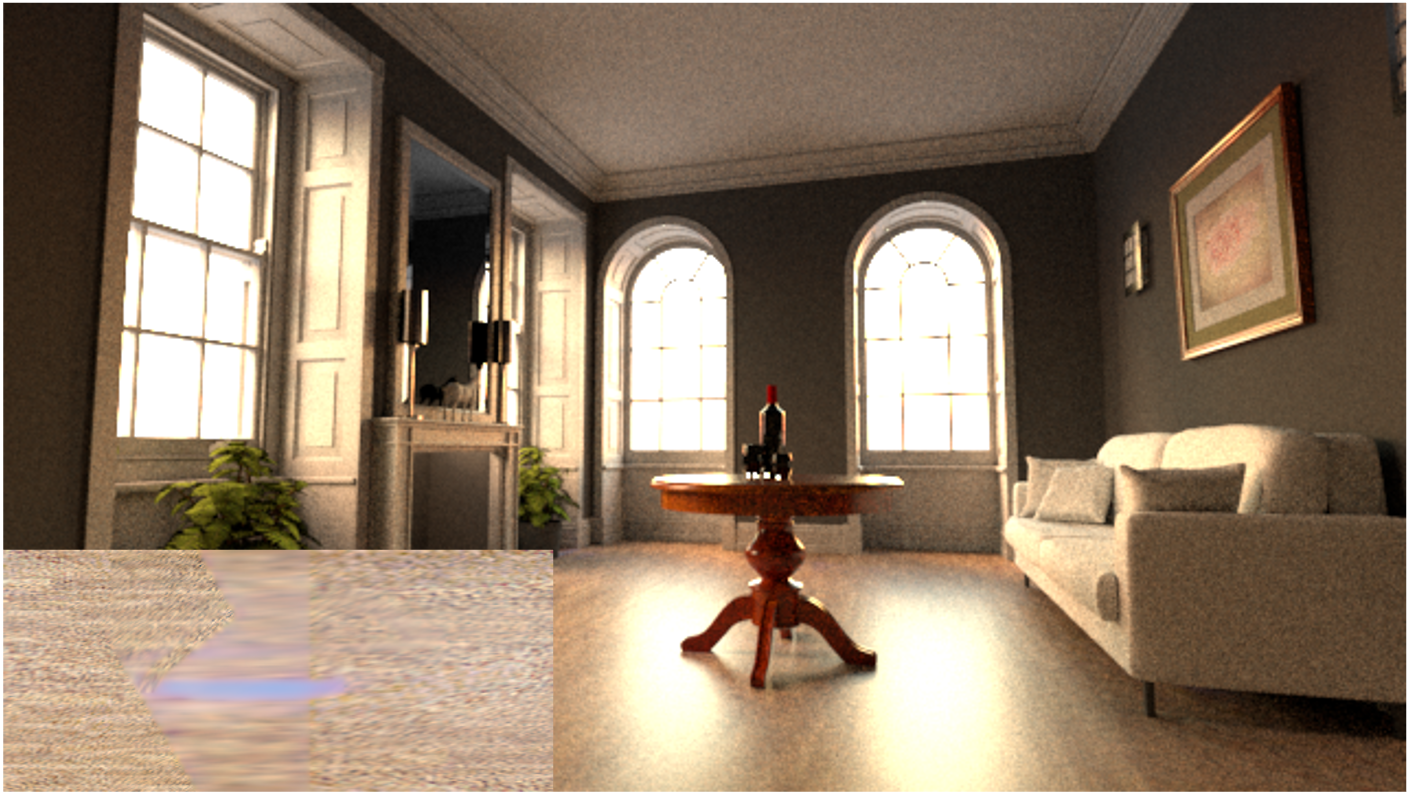} &
    \includegraphics[width=\mywidth\linewidth]{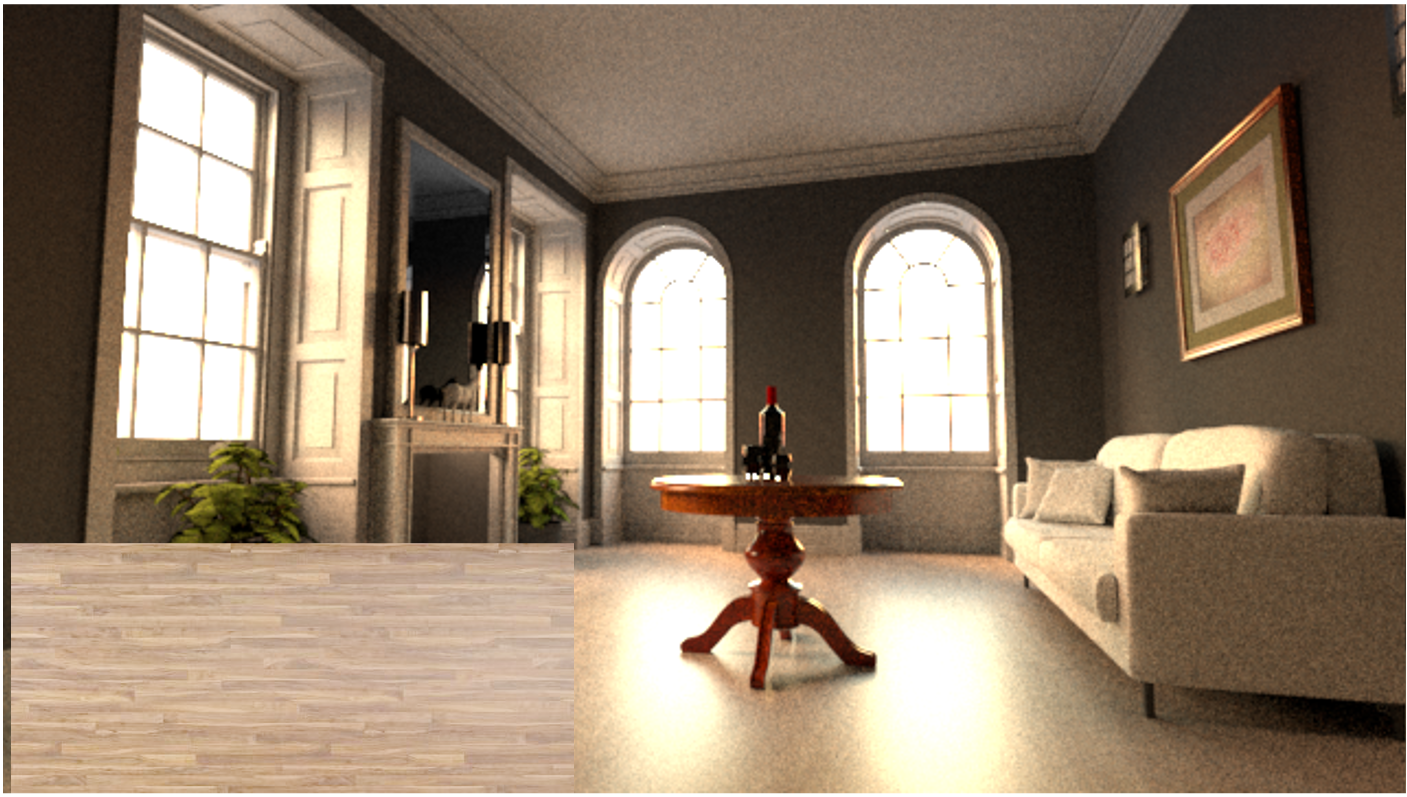} \\
    \includegraphics[width=\mywidth\linewidth]{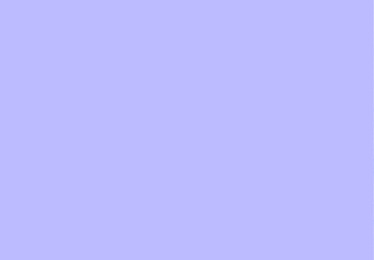} &
    \includegraphics[width=\mywidth\linewidth]{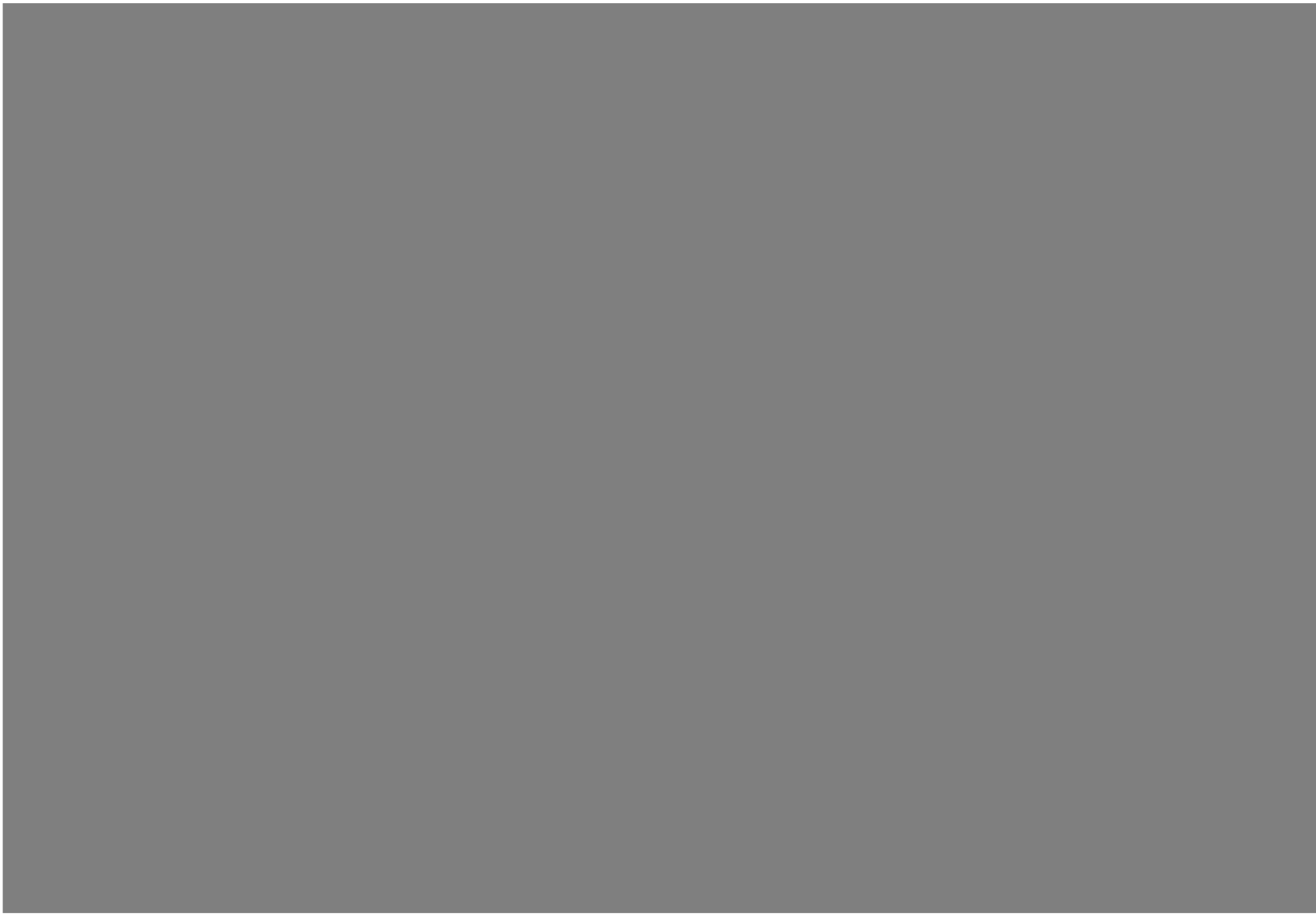} &
    \includegraphics[width=\mywidth\linewidth]{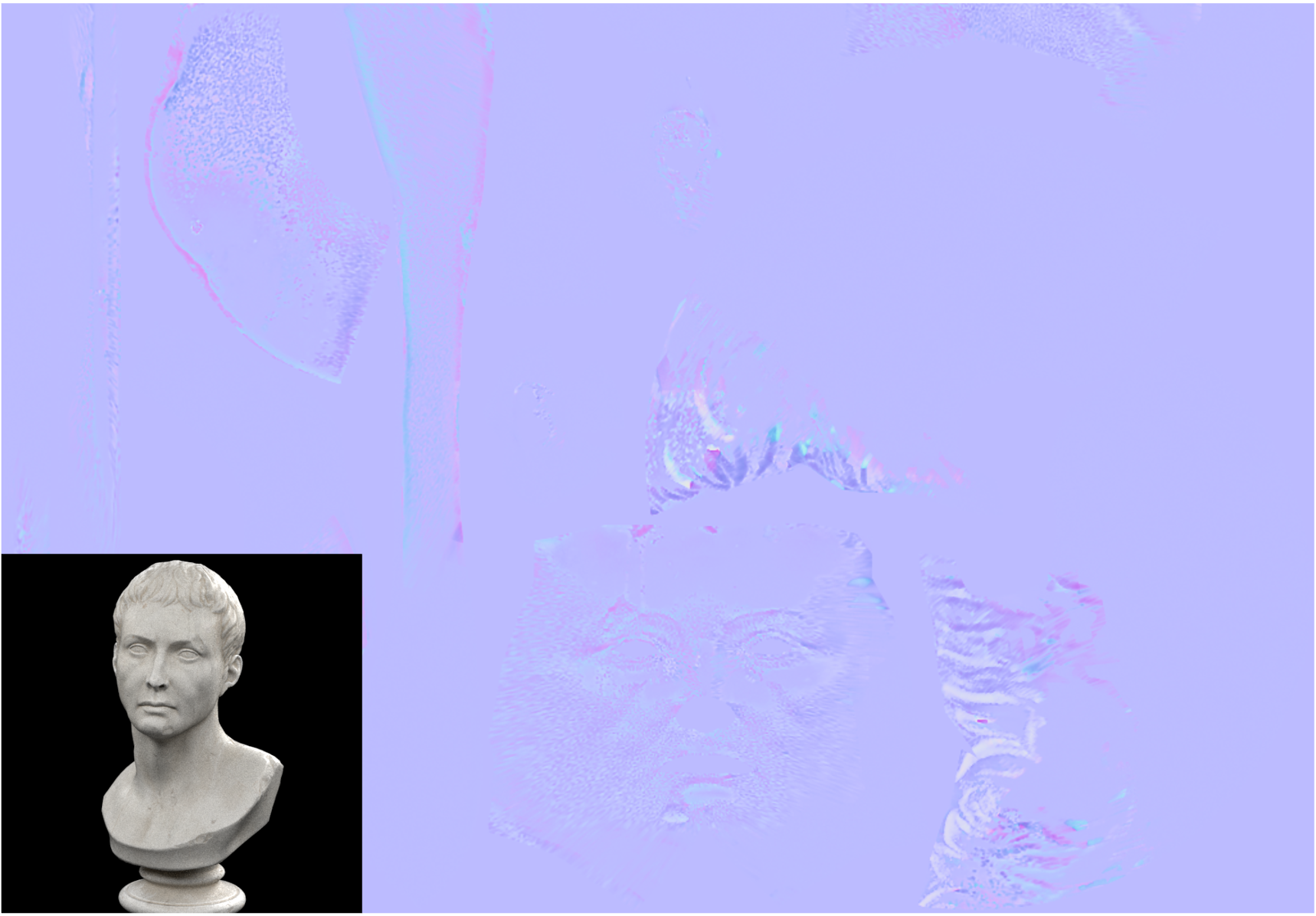} &
    \includegraphics[width=\mywidth\linewidth]{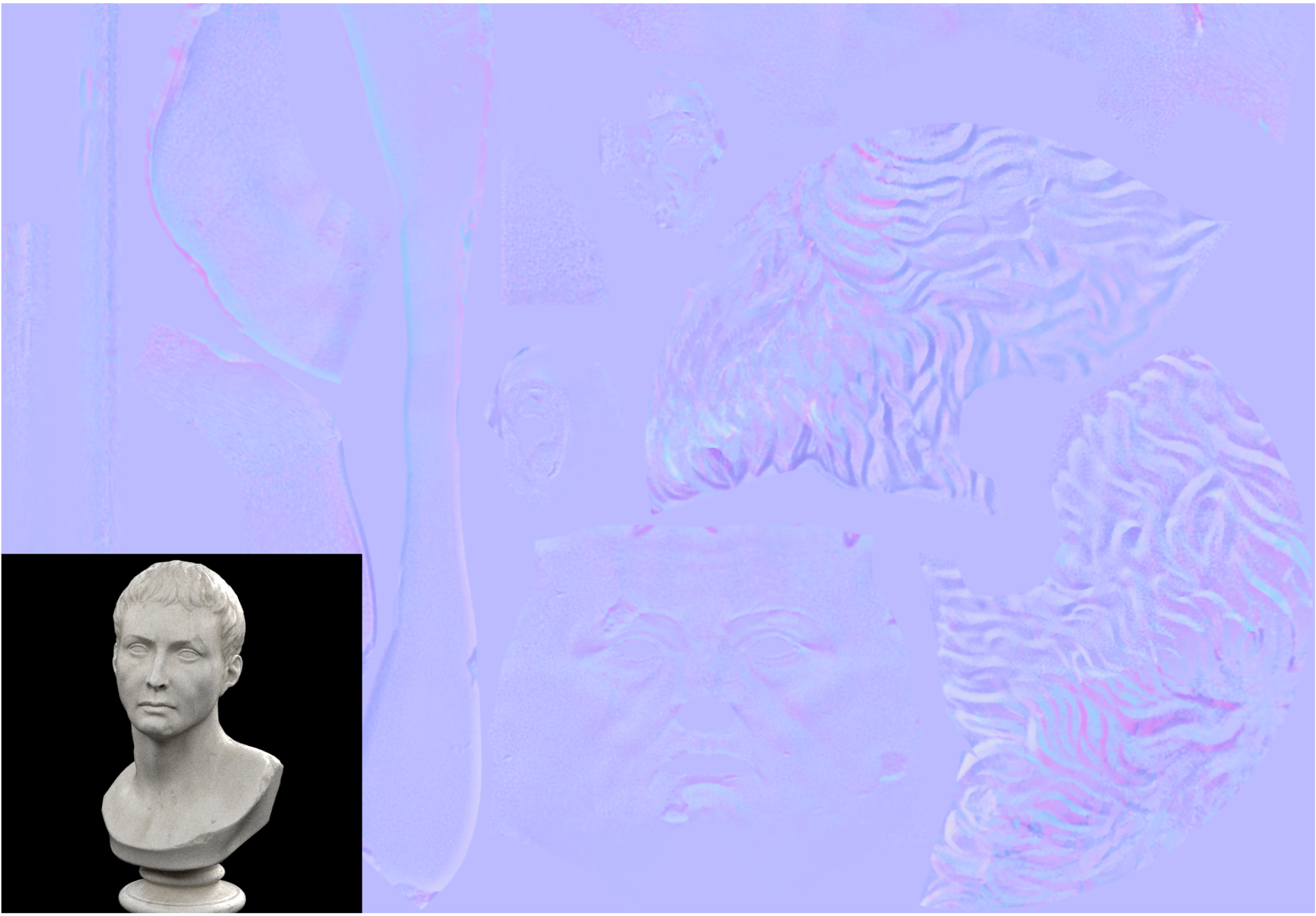} &
    \includegraphics[width=\mywidth\linewidth]{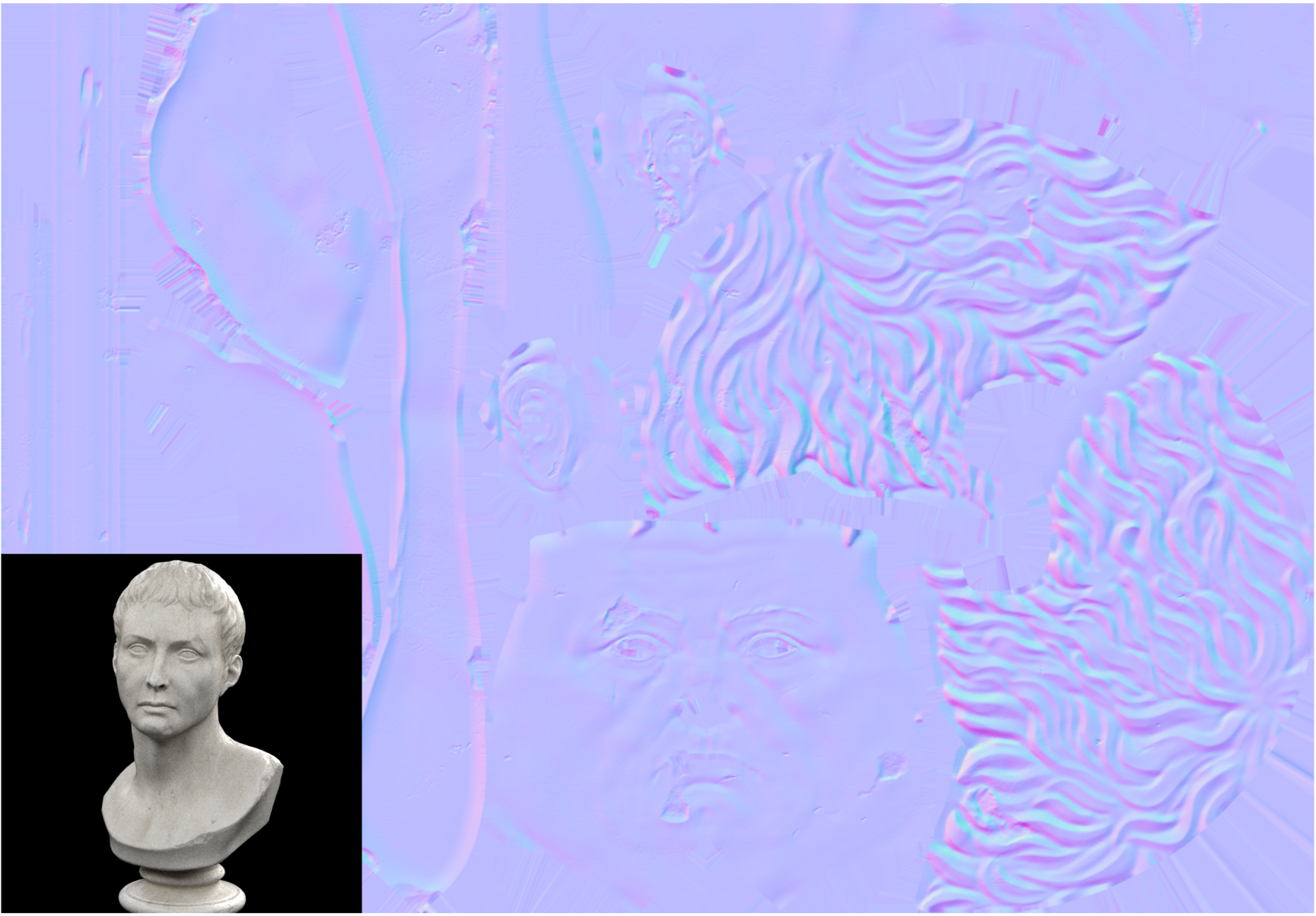} \\
};

\node[rmse] at (mymatrix-1-3.south east) {RMSE: 12.1848};
\node[rmse] at (mymatrix-1-4.south east) {RMSE: 3.955};
\node[rmse] at (mymatrix-2-2.south east) {RMSE: 0.1272};
\node[rmse] at (mymatrix-2-3.south east) {RMSE: 0.2491};
\node[rmse] at (mymatrix-2-4.south east) {RMSE: 0.1274};
\node[rmse] at (mymatrix-3-3.south east) {RMSE: 0.2679};
\node[rmse] at (mymatrix-3-4.south east) {RMSE: 0.0069};
\node[rmse] at (mymatrix-4-3.south east) {RMSE: 0.04761};
\node[rmse] at (mymatrix-4-4.south east) {RMSE: 0.04196};

\node[col_label] at (mymatrix-1-1.north) {Initial State};
\node[col_label] at (mymatrix-1-2.north) {Conv. AD};
\node[col_label] at (mymatrix-1-3.north) {PRB};
\node[col_label] at (mymatrix-1-4.north) {Ours};
\node[col_label] at (mymatrix-1-5.north) {Reference};

\node[row_label_left] at (mymatrix-1-1.west) {Kitchen};
\node[row_label_left] at (mymatrix-2-1.west) {Veach};
\node[row_label_left] at (mymatrix-3-1.west) {Living Room};
\node[row_label_left] at (mymatrix-4-1.west) {Marble Bust};

\node[row_label_right] at (mymatrix-1-5.east) {Polarizer $\theta$};
\node[row_label_right] at (mymatrix-2-5.east) {Diffuse + \\ Roughness};
\node[row_label_right] at (mymatrix-3-5.east) {Diffuse};
\node[row_label_right] at (mymatrix-4-5.east) {Normal Map};

\node[font=\sffamily\bfseries\Large, text=red] at (mymatrix-1-2.center) {OOM};
\node[font=\sffamily\bfseries\Large, text=red] at (mymatrix-3-2.center) {OOM};
\node[font=\sffamily\bfseries\Large, text=red] at (mymatrix-4-2.center) {OOM};

\end{tikzpicture}
\caption{\textbf{Polarized inverse rendering}. We evaluate four inverse rendering tasks (rows) and compare conventional automatic differentiation (Conv.~AD), unpolarized PRB~\cite{vicini2021PathReplay}, and our polarization-aware method against the ground truth reference. From left to right, we show the initial state, the optimized result using Conv.~AD, PRB, our method, and the reference. The tasks include optimizing the rotation angle of a linear polarizer (\emph{Kitchen}), jointly optimizing diffuse reflectance and roughness (\emph{Veach}), optimizing diffuse reflectance (\emph{Living Room}), and recovering a normal map (\emph{Marble Bust}). Conv.~AD fails to scale to several of these settings and runs out of memory (OOM). Compared to PRB, our method achieves lower reconstruction error (RMSE overlays) and produces results that more closely match the reference across all tasks.}
\label{fig:baseline_comparison}
\vspace{-0.5cm}
\end{figure*}

\cref{fig:scene_angle_lc_grid} shows reverse-mode gradients with respect to surface normals, roughness, and specular reflectance under these configurations. Each polarization state produces distinct gradient structures, indicating that polarization provides complementary information for parameter estimation.
Having verified both correctness and diversity of gradients, we next evaluate the practical impact of polarization in optimization tasks (\cref{fig:baseline_comparison}). We compare unpolarized PRB~\cite{vicini2021PathReplay} against our polarization-aware method.
In the first task (\textit{Kitchen} scene), we aim to remove glare in the cooking area (highlighted in red). We introduce a linear polarizer in front of the camera and optimize its orientation. Since PRB cannot differentiate parameters controlling the polarization state, it fails to suppress the glare. In contrast, our method successfully optimizes the configuration and reduces the specular reflection.

In the second task (\textit{Veach} scene, row~2), we jointly optimize the roughness and diffuse texture of a plate. PRB struggles to disentangle these parameters, leading to incorrect texture recovery (see inset). By leveraging cross and parallel polarized observations, our method better separates specular and diffuse components and recovers parameters closer to the reference. While this is a known application of polarization for reflectance separation \cite{ma_polarization,ghosh_multiview_polarization,Riviere2017}, it is important to note here that we do not explicitly pre-separate diffuse and specular observations to model with diffuse and specular BSDFs separately. Details of experimental design are in supplemental material. 

We observe similar behavior in the \emph{Living Room} scene when reconstructing the diffuse floor texture. Finally, we jointly optimize surface normals and diffuse reflectance of a marble bust from a multi-view dataset consisting of 12 images to recover a 1K-resolution normal map. PRB produces inaccurate normals, particularly in highly detailed regions such as the statue's hair. Polarization provides additional cues about surface orientation, enabling improved normal recovery.

Across these experiments, polarization enhances parameter identifiability, and our contribution is to enable such tasks with an efficient polarization-compatible integrator. While our results match Conv.~AD in correctness, AD does not scale to scenes with many bounces, large parameter spaces, or high resolutions; in several cases it runs out of memory (gray entries in \cref{fig:baseline_comparison}). 

\subsection{Geometry Reconstruction with Polarization}
\label{subsec:results_geometry}
\begin{figure}[t]
\centering
\setlength{\tabcolsep}{0pt}
\renewcommand{\arraystretch}{0}
\begin{tabular}{@{}c@{}c@{}c@{}c@{}c@{}c@{}}
Input & Pol & Unpol & {} Test View & Pol & Unpol \\

\includegraphics[scale=0.15]{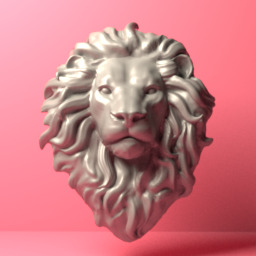} &
\includegraphics[scale=0.15]{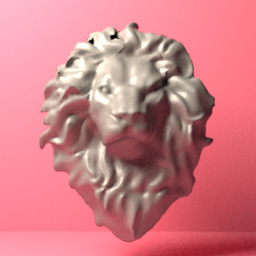} &
\includegraphics[scale=0.15]{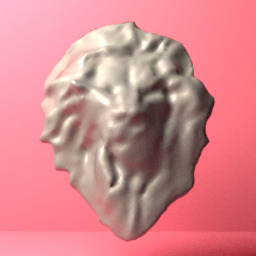} &
{}
\includegraphics[scale=0.15]{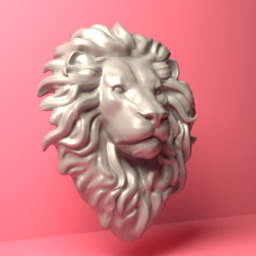} &
\includegraphics[scale=0.15]{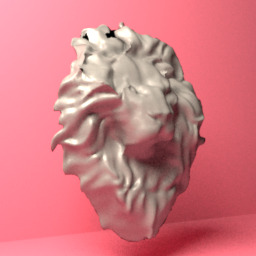} &
\includegraphics[scale=0.15]{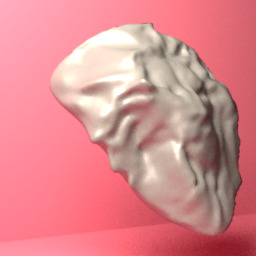} \\

\includegraphics[scale=0.15]{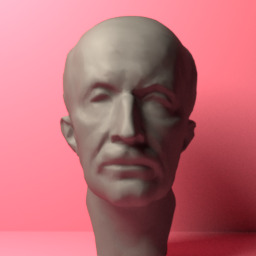} &
\includegraphics[scale=0.15]{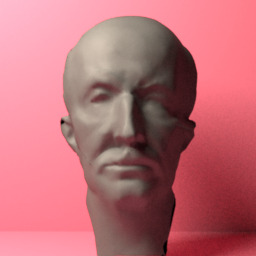} &
\includegraphics[scale=0.15]{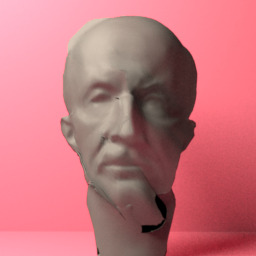} &
{}
\includegraphics[scale=0.15]{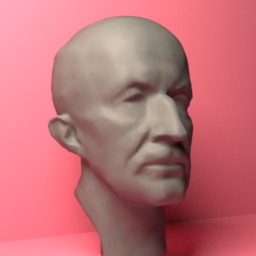} &
\includegraphics[scale=0.15]{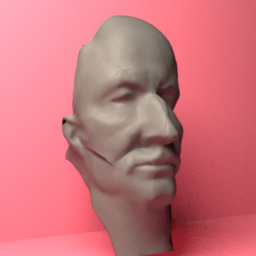} &
\includegraphics[scale=0.15]{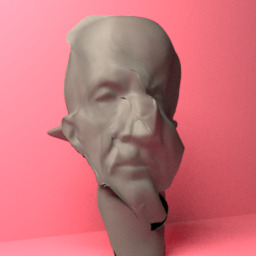} \\

\includegraphics[scale=0.15]{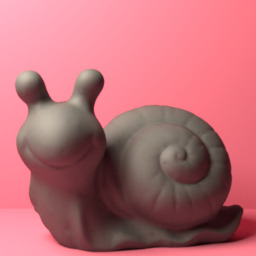} &
\includegraphics[scale=0.15]{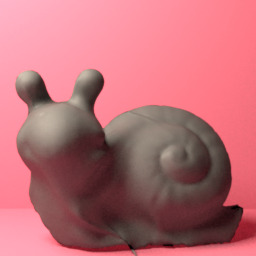} &
\includegraphics[scale=0.15]{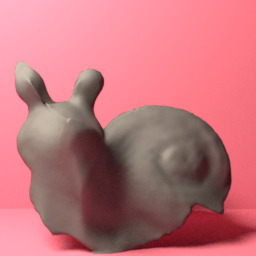} &
{}
\includegraphics[scale=0.15]{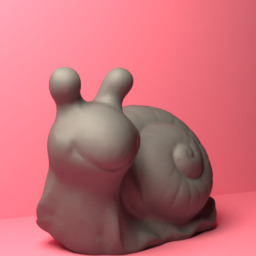} &
\includegraphics[scale=0.15]{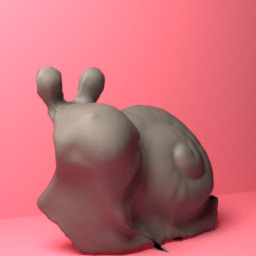} &
\includegraphics[scale=0.15]{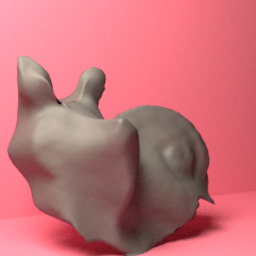}

\end{tabular}
\vspace{-0.3cm}
\caption{
Single-view 3D reconstruction. Ours (\textit{Pol.}) reconstructed geometry is sharper than the unpolarized baseline (\textit{Unpol.}), shown from  test views.
}
\label{fig:3d_recon}
\vspace{-0.3cm}
\end{figure}

Classical shape-from-polarization methods exploit the fact that the polarization state of reflected light carries surface-orientation information. To demonstrate this, we further extend the projective sampling integrator \cite{Zhang2023Projective} to polarized light transport with suffix caching, enabling visibility-discontinuity gradients while retaining our inversion-free polarized replay.
We evaluate single-view 3D reconstruction over seven meshes, comparing our polarized variant against the corresponding unpolarized projective-sampling baseline under the same initialization and optimization schedule. As shown in \cref{fig:3d_recon}, polarized observations recover sharper geometric details and more faithful silhouettes that are smoothed or distorted in the unpolarized reconstructions. Quantitatively, polarization reduces Chamfer distance from $\mathbf{0.1874}$ to $\mathbf{0.0904}$ and normal error from $\mathbf{65.61}^{\circ}$ to $\mathbf{46.72}^{\circ}$. These results indicate that polarized light transport provides crucial shape cues beyond intensity alone, and that our replay formulation can be combined with visibility-aware shape-gradient estimators. We analyze scalability and performance in the next section.

\subsection{Performance Evaluation}
\label{subsec:results_benchmark}
\begin{figure*}[t]
    \centering
    \newcommand{\subfigwidth}{0.24\textwidth}

    \begin{subfigure}[t]{\subfigwidth}
        \centering
        \includegraphics[width=\linewidth]{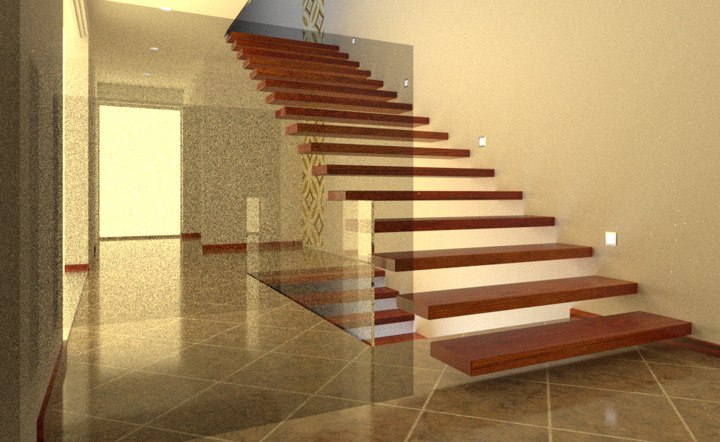}
        \caption{Reference Scene}
        \label{fig:bench_scene}
    \end{subfigure}
    \hfill 
    \begin{subfigure}[t]{\subfigwidth}
        \centering
        \includegraphics[width=\linewidth]{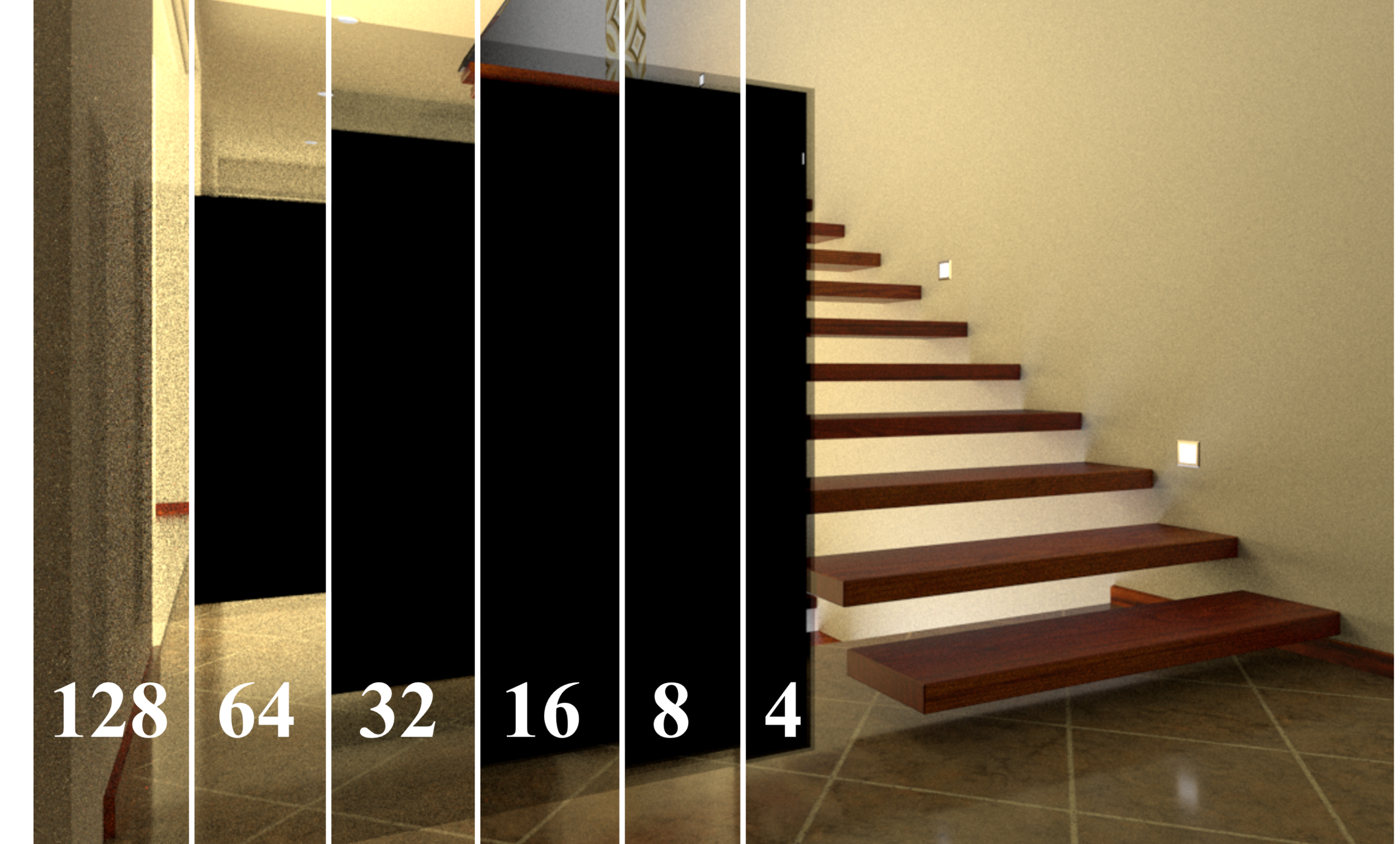}
        \caption{with N bounces}
        \label{fig:bench_metric}
    \end{subfigure}
    \hfill 
    \begin{subfigure}[t]{\subfigwidth}
        \centering
        \includegraphics[width=\linewidth]{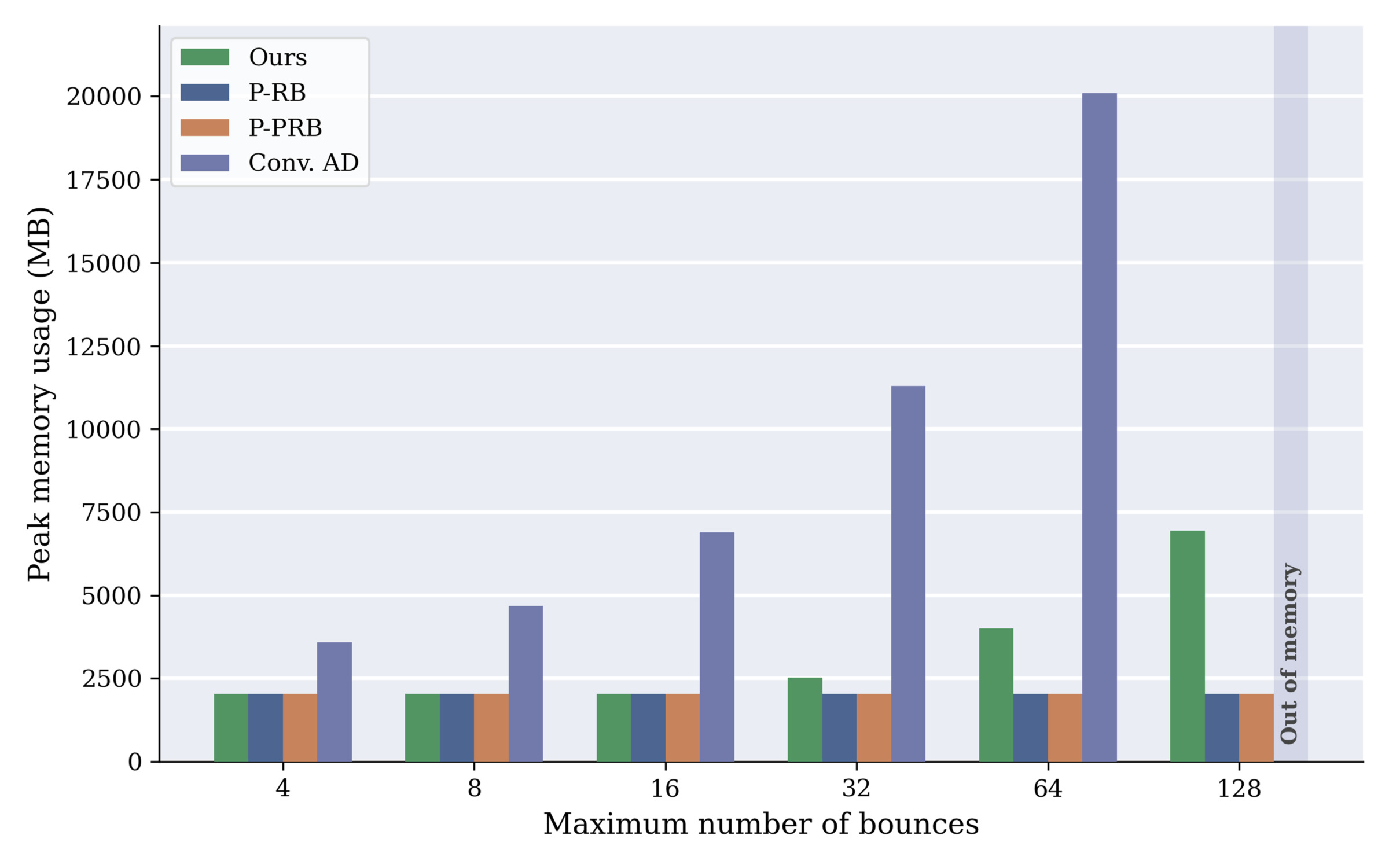}
        \caption{Memory Usage}
        \label{fig:bench_memory}
    \end{subfigure}
    \hfill 
    \begin{subfigure}[t]{\subfigwidth}
        \centering
        \includegraphics[width=\linewidth]{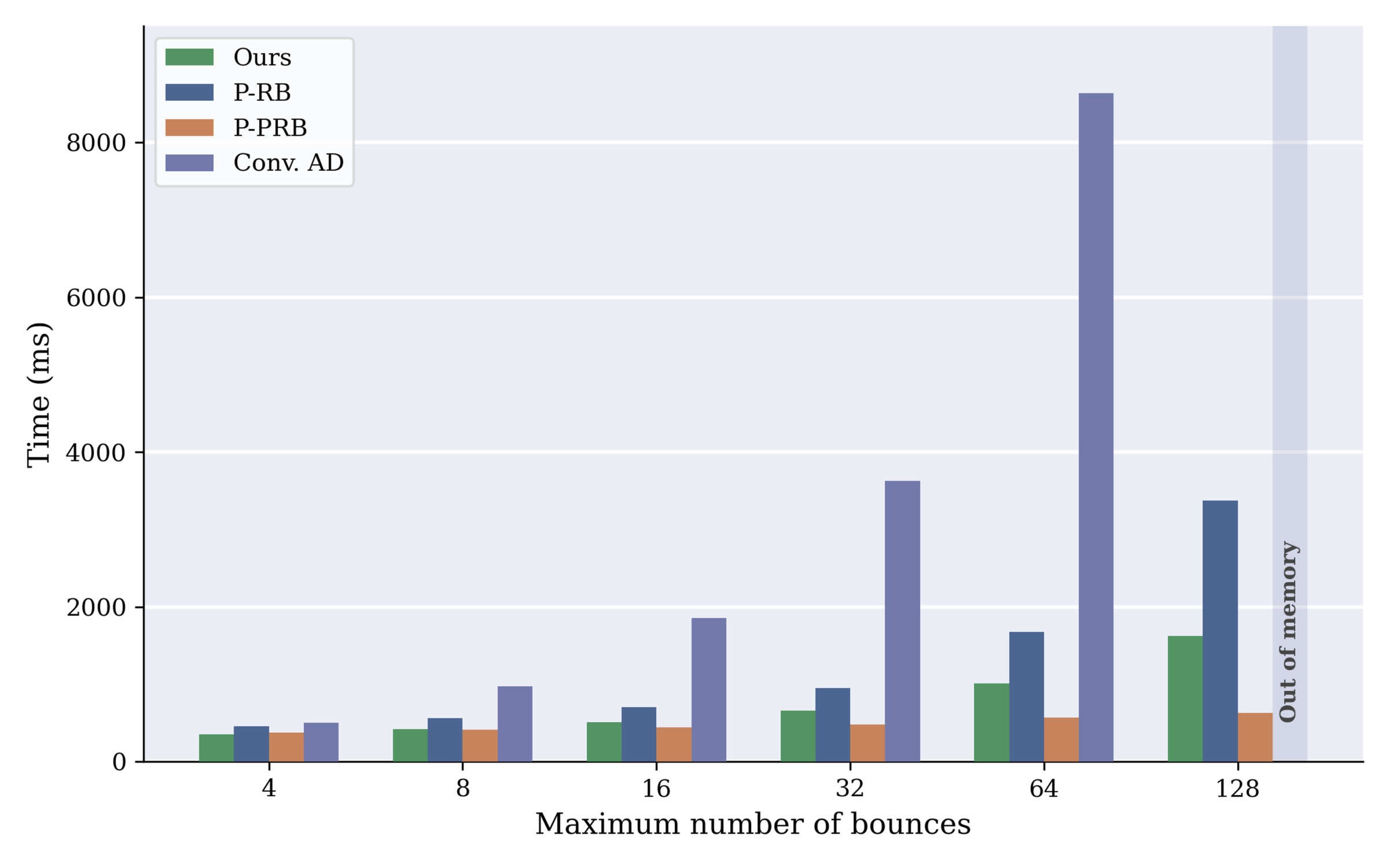}
        \caption{Time}
        \label{fig:bench_runtime}
    \end{subfigure}

    \caption{Performance evaluation on a polarized scene rendered at $1280\times720$ resolution with $1$ sample per pixel.
    (a) We use the \emph{Modern Hall} scene and introduce a stack of 85 linear polarizers to simulate long light paths.
    (b) We illustrate the effect of increasing the maximum path depth on the rendered appearance.
    We then compute gradients with respect to the floor albedo using conventional automatic differentiation (Conv.~AD), P-RB, P-PRB, and our method.
    (c) Conv.~AD exhausts GPU memory of a NVIDIA GeForce RTX 5090, whereas our method maintains an approximately constant memory up to depth 32 and increases only slightly thereafter, while remaining substantially below Conv.~AD.
    (d) P-PRB attains low runtime but produces inaccurate gradients. P-RB produces unbiased gradients, but its runtime is higher; in this experiment it is approximately $2\times$ slower than our method.}
    \label{fig:benchmark_time_memory}
    \vspace{-0.5cm}
\end{figure*}

Mitsuba represents polarized quantities using per-channel Mueller matrices. Instead of a simple 3D vector, each radiance quantity becomes a $4 \times 4 \times 3$ tensor. This increased representation cost makes differentiable rendering in polarized mode significantly more demanding in both memory and computation time, motivating the need for an efficient gradient estimator.

We benchmark reverse-mode gradient computation in terms of runtime and memory consumption (\cref{fig:benchmark_time_memory}). Thanks to the use of local memory, our method exhibits nearly constant memory usage up to a path depth of 32, with only a slight increase beyond that point. It always remains substantially more memory-efficient than conventional AD. In the supplemental material, we discuss a variant of our method that combines local caching with recursive recomputation based on polarization heuristics to further reduce memory consumption.
We also evaluate the hybrid variant introduced in \cref{subsec:method_hybrid}. For the path depths used in our main applications, the fully cached method remains the most efficient choice. The hybrid variant becomes useful at larger depths, where storing suffix values at every interaction increases local memory use. 
%
%
Detailed plots are included in the supplemental material.

Although P-PRB also demonstrates constant memory behavior, its gradient inaccuracies limit its applicability in polarized settings. Conventional AD, while producing correct gradients, incurs substantially higher memory and runtime costs. In comparison, our method achieves correct gradients with improved efficiency, and is approximately twice as fast as P-RB in our experiments, making it a practical approach for polarized differentiable rendering.

\section{Conclusion and Future Work}

In this paper, we introduced a memory-efficient differentiable path tracing method for polarized light transport. We showed that standard PRB does not directly generalize to polarimetric rendering, since common Mueller operators can be rank-deficient and make replay-based inversion ill-defined. Our method avoids this issue using cached suffix replay, and further introduces a hybrid cache--recompute variant for larger path depths. Together, these algorithms enable stable and efficient gradient computation for polarized scenes. We validated the method through gradient comparisons against conventional AD and demonstrated that polarimetric measurements improve material, texture, and normal recovery compared to unpolarized baselines.

Future work includes extending the framework to subsurface scattering. Since multiple scattering tends to depolarize light while the first surface interactions can retain polarization cues, differentiable polarized subsurface transport could help separate surface specularities from internal scattering in translucent materials such as marble, jade, or skin. It would also be interesting to apply our method to computational design problems such as waveguide design~\cite{yang2026end}.

\section*{Acknowledgements}

This work was supported by the ERC Consolidator Grant 4DRepLy (770784) and the Saarbrücken Research Center for Visual Computing, Interaction and AI. We thank Peter Kultis for helpful discussions, the anonymous reviewers for their constructive feedback, and Shrisha Bharadwaj for insightful discussions, careful proofreading, and invaluable support throughout the project.

\clearpage

\bibliographystyle{splncs04}
\bibliography{main}
\end{document}